# A novel approach to data generation in generative model

JaeHong Kim[1, 2], Jaewon Shim[3]

[1] Healthcare, Legal and Policy Center, Graduate school of Law, Korea University, Seoul 02841, Korea

[2] Human-Inspired AI Research, Korea University, Seoul 02841, Korea

[3] Center for 0D Nanofluidics, Institute of Applied Physics, Department of Physics and Astronomy, Seoul National University, Seoul 08826, Korea

Contact email: comforta@korea.ac.kr

## Abstract

Variational Autoencoders (VAEs) and other generative models are widely employed in artificial intelligence to synthesize new data. However, current approaches rely on Euclidean geometric assumptions and statistical approximations that fail to capture the structured and emergent nature of data generation. This paper introduces the Convergent Fusion Paradigm (CFP) theory, a novel geometric framework that redefines data generation by integrating dimensional expansion accompanied by qualitative transformation. By modifying the latent space geometry to interact with emergent high-dimensional structures, CFP theory addresses key challenges such as identifiability issues and unintended artifacts like hallucinations in Large Language Models (LLMs). CFP theory is based on two key conceptual hypotheses that redefine how generative models structure relationships between data and algorithms. Through the lens of CFP theory, we critically examine existing metric-learning approaches. CFP theory advances this perspective by introducing time-reversed metric embeddings and structural convergence mechanisms, leading to a novel geometric approach that better accounts for data generation as a structured epistemic process. Beyond its computational implications, CFP theory provides philosophical insights into the ontological underpinnings of data generation. By offering a systematic framework for high-dimensional learning dynamics, CFP theory contributes to establishing a theoretical foundation for understanding the data-relationship structures in AI. Finally, future research in CFP theory will be led to its implications for fully realizing qualitative transformations, introducing the potential of Hilbert space in generative modeling.

**Keywords:** Variational Autoencoders (VAEs), Convergent Fusion Paradigm (CFP), Data Manifold Hypothesis, Metric Learning, Artificial Intelligence and Ontology,

# I. Introduction

In the field of artificial intelligence, autoencoders (AEs) are generally understood as operating through a two-step process: data compression and reconstruction (Hinton and Salakhutdinov 2006). Variational Autoencoders (VAEs) extend this approach by introducing a variational distribution $q(z|x)$ in place of the posterior distribution $p(z|x)$, which is ultimately replaced by the prior distribution $p(z)$ (Bishop 2006). This three-step process—compression, transformation, and reconstruction (or generation)—is based on the fundamental understanding that VAEs have evolved from AEs. Moreover, VAEs benefit from mathematical approximations that facilitate efficient computation (Cybenko 1989; Hornik et al. 1989). However, this approach oversimplifies the process of data generation, which is distinct from compression and reconstruction. The conventional VAE framework risks distorting the actual process of data generation, often leading to unintended results. This issue affects a broad range of generative AI models, including Large Language Models (LLMs), which frequently exhibit hallucinations—outputs that are not grounded in the given data (Mittelstadt et al. 2023). While part of this issue arises from mathematical descriptive constraints, we argue that a more fundamental problem is the lack of a rigorous data science theory that can systematically abstract and idealize the process of data generation.

To address this gap, this paper introduces the Convergent Fusion Paradigm (CFP) as a philosophical foundation for data science theory that provides a novel geometric perspective on data generation within generative models such as VAEs. The CFP theory constructs a data geometry that enables a structured integration of low- and high-dimensional data spaces, ensuring seamless convergence across different dimensional levels. By applying this framework to generative models such as VAEs, high-level information embedded in high-dimensional data can be effectively captured without disrupting the intricate interplay between low- and high-dimensional data structures. Specifically, CFP theory enables VAEs to model data generation more precisely by fusing the high-dimensional ambient input space with the low-dimensional latent space via a Riemannian metric, thereby yielding a new high-dimensional ambient output space.

This study argues that a novel modeling approach for data-algorithm relationships in Deep Neural

Networks (DNNs)—distinct from traditional statistical modeling—is required. To establish this perspective, we propose CFP theory as a new geometric framework for data science, built upon two conceptual hypotheses that describe a process of dimensional expansion accompanied by qualitative transformation **(Sections 2-1, 2)**. Furthermore, applying CFP theory to existing generative models offers new insights into the data manifold hypothesis, particularly in relation to the Riemannian metric **(Section 2-3)**. Next, this study critically examines the work of (Arvanitidis et al. 2018, 2020, 2021), which explores how to define an appropriate Riemannian metric for the structural spaces of VAEs, specifically the ambient space and latent space. These studies seek to establish an appropriate structure for the ambient space and latent space, with the aim of deriving a Riemannian metric that minimizes the distance between data points, thereby producing the optimal outcome through VAEs. This paper summarizes and evaluates their approaches and investigates how their methods can be reinterpreted within the framework of CFP theory to enhance their theoretical and practical contributions **(Section 3)**. Finally, this study outlines key research challenges from the structural standpoint of CFP theory **(Section 4-1)** and proposes a philosophical perspective—a novel area by Passivity of CFP (Leave as It is, LAI)—as a conceptual foundation for addressing these challenges **(Section 4-2)**.

## II. CFP Theory to provide a novel data geometry for generative model

### 1. Background: Modeling relationship between Data and Algorithms in DNNs

To understand CFP as a novel theoretical approach that provides a data geometric framework for capturing the process of data generation in generative models, it is essential to examine the fundamental difference between DNNs and Euclidean geometry. The former represents an open-world data science paradigm, whereas the latter is based on a closed-world assumption. Euclidean geometry assumes a closed and self-contained space that seeks uniqueness and completeness through axioms. This assumption restricts the representation of space and time to a limited coordinate system (Greenberg 1993). Consequently, dimensional expansion in Euclidean space is often interpreted merely as a quantitative addition within a two-dimensional plane. In contrast, DNNs, which assume an open world, seek to implement a performance that achieves the correlation between the data, representing the facet of human life, and the algorithm, encapsulating the intended objectives through the expansion of dimensions embedded in the data features.

Based on this premise, it is crucial to recognize that the modeling of DNNs differs from the two statistical modeling cultures described by (Breiman 2001): data modeling culture and algorithmic modeling culture. Unlike traditional data statistics, which either derives specific algorithmic models from data (data modeling culture) or seeks an algorithm that predicts data irrespective of its inherent structure (algorithmic modeling culture), DNNs must undergo a dual process. This involves first identifying a learning algorithm through training data and subsequently determining a model by fitting it to domain data (LeCun et al. 2015).[1] In other words, if the core components of a DNN are tasks, models, and features, it is necessary to distinguish between the process of learning problems that generate models and the task execution performed by models. As illustrated in Figure 1, the process of modeling, created through the interaction between learning algorithms and data, is not merely a set of parameters (of, for example, a classifier) defined by data features only (Flach 2012). In other words, tasks are addressed by models, whereas learning problems are solved by learning algorithms that produce models (Flach 2012).

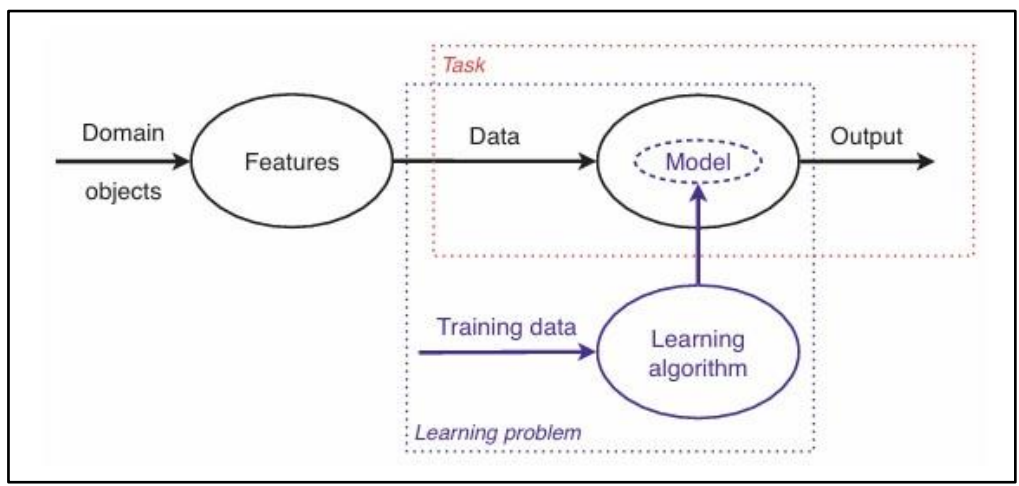

Figure 1. "Data-Algorithm Interrelationship Modeling" (Image adapted from (Flach 2012))

Despite this distinction, many AI researchers fail to recognize the difference between the formation of learning algorithms in the "learning problem" domain (the purple dotted area in Figure 1) and the construction of models in the "task" domain (the red dotted area in Figure 1). While both might appear to follow a fitting-based approach in the case of linear regression, their content fundamentally differs due to the dual nature of the process. The combination of a learning algorithm derived from training data and domain data results in outputs that are inherently distinct. As previously mentioned, the model in Figure 1 is neither solely defined by domain data features nor merely a set of parameters specified by the learning algorithm.

Taking this further, the correlation between data and learning algorithms, referred to as the "model" in Figure 1, becomes increasingly complex as the number of nodes in a DNNs increases exponentially. This complexity leads to the emergence of activation in hidden units within these intricate interrelationships (Bau et al. 2018; Boge 2023; Woodward 2005). Methods such as linear regression, which operate post hoc and within a one-dimensional framework, can only measure linear changes in models formed by combining domain data with algorithms trained from data. However, as the dual-process-based workspace allows for the expansion of increasingly complex node structures,

---

[1] The process of fitting a learning algorithm derived from training data to domain data is classified into supervised learning, unsupervised learning, and reinforcement learning.

unanticipated synergistic effects can emerge (López-Rubio 2021).

Considering the structure of DNNs, it is more appropriate to focus on the relationship between data and algorithms rather than viewing DNN modeling as a conflict between data and algorithm modeling cultures, as proposed in traditional data statistics (Breiman 2001). In this context, this study defines "modeling the relationship between data and algorithms" as a unique modeling approach in DNNs and aims to theoretically examine the new developmental phase introduced by this distinctive form of DNN modeling. This theoretical exploration is expected to establish a new framework that goes beyond the conventional assumptions embedded in existing mathematical and logical methodologies.

## 2. Expansion of Dimensions Accompanied by Qualitative Changes – Two New Conceptual Hypotheses

This study proposes that the new developmental phase created by DNNs' unique "modeling of the relationships between data and algorithms" is not a quantitative expansion of dimensions through Euclidean geometry as previously discussed, but rather an "expansion of dimensions accompanied by qualitative transformations." In other words, this study argues that the development of the relationship between data and algorithms in data science is not merely a quantitative scaling-up of dimensions[2] but can be better understood—figuratively speaking—as akin to a *'plane'* undergoing a qualitative transformation to become a *'solid figure.'* Particularly in the case of generative artificial intelligence (Goodfellow et al. 2014; Kingma and Welling 2013; Rombach et al. 2022), which goes beyond the level of implementing simple machinery mechanisms,[3] the developmental phase of the relationship between data and models—in other words, the "dimensional expansion accompanied by qualitative transformations"—should not be overlooked. This developmental process organizes high-dimensional structures as emergent constructs, distinct from the lower-dimensional structures, at the level of data features (Capra and Luisi 2014).[4]

This raises two fundamental questions: First, what it means, in concrete terms for the relational development between data and algorithms in data science, to embody a sort of transformation of planes into solid figures, accompanied by a qualitative transition, rather than to simply scale up in a quantitative manner. In sequence, it highlights the challenge of determining how mathematical tools, traditionally more suitable for closed systems, can appropriately and accurately encapsulate the developmental

---

[2] Therefore, mathematically, it can be represented on a coordinate plane within Euclidean space.

[3] Early linear regression analysis follows this approach. However, Linear Regression Analysis (LRA) still constitutes a fundamental structure in most DNNs and should not be treated lightly.

[4] It is important to note that due to the influence of reductionism, the high-dimensional structure of data should not be understood merely as the aggregation of lower-dimensional structures. Understanding data within a lower-dimensional framework differs fundamentally from understanding it within a high-dimensional structure. In this context, the high-dimensional structure of data should be regarded as an emergent, novel construct distinct from the lower-dimensional structure.

trajectory of DNNs as a field of data science that aims towards an open-world paradigm.

To address these issues, this study approaches the meaning of "dimensional expansion accompanied by qualitative transformation" through two conceptual guiding hypotheses[5]:

1. The concept of "Creating Relative Space-Time in Relationship (crSTR)" – This hypothesis explores the process by which a lower-dimensional entity transforms into a higher-dimensional framework, analyzing this transformation through a relational perspective. This concept may also be extended and substituted with "Another Bigger I Converging Newly in Days (ABICND)," signifying an entity that continuously evolves towards broader relationships.

2. The concept of "Duplex Contradictory Paradoxical Stratified structures of Thorough Closure (Solitude) – Eternal Opening (DCPSs of TC-EO)" – This hypothesis investigates the underlying force that drives dimensional expansion, serving as the catalyst that facilitates qualitative transformation into higher-dimensional structures. It suggests that this transformation emerges through "Thorough Closure" as a necessary precursor to "Eternal Opening."

However, due to the scope limitations of this study, an in-depth exploration of these concepts will be addressed in a forthcoming work titled "Philosophical Basis of Data-Algorithm Relationships Model". In this study, the focus will be placed primarily on the first conceptual hypothesis, constructing its logical framework based on foundations from physics and biology. Regarding the second hypothesis, a brief examination will be conducted in relation to Hofstadter's (1999) renowned work, Gödel, Escher, Bach: An Eternal Golden Braid (GEB), (Hofstadter 1999), particularly its discussion on the concept of the "strange loop." This examination will suggest that the second conceptual hypothesis may serve as a structural framework required when transforming from one dimensional level to a higher one.[6]

**1) First Conceptual Hypothesis – Creating Relative Space-Time in Relationships (crSTR)**

---

[5] Personally, I believe that properly designing this aspect can resolve many challenges in AI and open the door to surpassing the limitations of traditional data statistical modeling. These two conceptual hypotheses originate from the intersubjectivity of beings, forming the foundation of relationship philosophy. Relationship philosophy, in this context, is rooted in the classical Confucian text Zhongyong (中庸), as well as the later French existentialist philosophy of Henry Bergson and Emmanuel Levinas. This framework serves as the basis for the social science research model known as the Convergent Paradigm (CP). CP theorizes that true social harmony and prosperity emerge not merely through calculations of mutual gains and losses between independent entities but through the continuous acquisition of new identities beyond individual subjects via organic relational formations.

[6] For further discussion on this topic, refer to the author's other paper, A New Approach to Relationships Between Data Features: A Comprehensive Examination of Mathematical, Technological, and Causal Methodologies (Kim 2025).

To capture the transition of a 'plane' undergoing qualitative transformation into a 'solid figure,' it is necessary to establish the concept of "Creating Relative Space-Time in Relationship (crSTR)." This requires an understanding of the concept of time as explored in various scientific paradigms.

**A. Meaning of Time in Classical Newtonian Mechanics (Barbour 1982; Cohen and Smith 2002)**

In Newtonian or Galilean geometry, time is understood as an absolute coordinate system used to describe events. Regardless of where an event occurs, observers interpret and experience it identically. For instance, if one knows the initial state of a particle, they can predict its entire trajectory until its final state. In this classical approach, space is treated as a fixed entity, and time is merely a continuous parameter used for measurement.

**B. Meaning of Time in Einstein's Theory of Relativity (Hartle 2021)**

Einstein's theory of relativity introduced a fundamental shift in the understanding of time. Unlike in classical mechanics, events are no longer absolute but depend on the state of the observer. The same event can appear simultaneous to one observer but occur at different times to another due to variations in gravitational influence. Einstein incorporated time as a coordinate axis in his equations, treating it alongside space within a four-dimensional tensor. This approach fundamentally reframes time as a geometric dimension rather than an independent parameter. Consequently, an entity's inherent state is perceived in terms of its energy rather than time itself.

**C. Meaning of Time in Quantum Mechanics (Dalmonte and Montangero 2016; Kogut 1983)**

Quantum mechanics provides an alternative perspective on time, treating it similarly to a wave-like field that emerges from the interaction of objects. According to Heisenberg's uncertainty principle, $\Delta x \Delta p \geq \hbar/2$, when rewritten in terms of energy and time, it states that $\Delta E \Delta t \geq \hbar/2$. This implies that energy and time cannot be simultaneously determined with precision—an increase in accuracy for one lead to a loss of precision in the other. As a result, quantum mechanics views time as a discontinuous wave field. Furthermore, in the microscopic world governed by quantum mechanics, time is structured as a discrete mesh, reinforcing the concept of time as a field rather than an absolute continuum.

**D. Summary – Hypothesis of Dimensional Expansion embodying crSTR or ABICND**

If time is inseparable from space and must be understood as a field-like, discontinuous wave, then the concept of time can be theoretically formulated within relational contexts.[7] To illustrate this, consider the chemical reaction in which hydrogen and oxygen molecules combine to form water. This reaction represents not merely a numerical aggregation but the formation of a directional field of existence. Once this reaction occurs, the relationship between the elements generates a new form of directed energy (Bernath 2020). Similarly, the creation of relative space-time within relationships (crSTR) suggests that dimensional expansion does not occur in isolation but emerges from the inherent interconnectivity of entities, reinforcing the hypothesis of continuous relational evolution toward higher dimensions.

---

[7] It is crucial to note that defining time alongside space does not necessarily mean that time must always be spatialized, as suggested in Einstein's interpretation.

To elucidate this phenomenon in a manner more readily comprehensible to those versed in terms of physical sciences, it can be described as follows: it is described that when two chemicals meet and their electron affinities are aligned, a chemical bond is formed, resulting in the creation of a new substance. In the context of physics, the concept of "electron affinity" is understood as the eigenfrequency of a newly generated substance. When the eigenfrequencies of two substances match (resonate), a chemical bond is formed instantaneously at the optimal distance for energy transfer, resulting in the creation of a new substance. The eigenfrequency can be understood as the unique time-scale of each substance (Bernath 2020). The new substance that is created by the unique time-scale of each substance coming together, along with the other unique time-scale that is identified within it, can be inferred as created relative Space-Time in Relationship (crSTR). As this crSTR aims to create a new existence, it can also be identified as the creation of a new substance (Auffray et al. 2003; Auffray and Nottale 2008).

It is challenging to encapsulate the aforementioned concepts within the confines of mathematical and physical axioms, particularly given the prevailing constraints of scientific rigidity. Nevertheless, it is possible to construct it as a theory based on this hypothesis. If such a theory does not contravene the established mathematical and physical standards, it will not present an inherent issue. Furthermore, there is no issue in making use of the theory by establishing the modelling technique of DNN as one of the organizational theories of AI. This is due to the fact that the development process of AI has been carried out by incorporating ideas from the development of various disciplines, including symbolism derived from logic (Domingos 2015), connectionism motivated by neuroscience (Baek et al. 2021), and so forth.

Accordingly, further philosophical elucidation may be required. However, if we organize the above-mentioned hypothesis, it can be stated as follows:

$$A \frown B \Longrightarrow crSTR \fallingdotseq ABICND$$

When two entities, A and B, exist within the same dimension and establish a relationship, they create a relative Space-Time in Relationship (crSTR) (①).

Furthermore, the relative Space-Time acts as a field (場) that constructs a higher-dimensional framework, ultimately converging towards a new entity known as "Another Bigger I Converging Newly in Days (ABICND)" (②).

Table 1. New Conceptual hypothesis for dimensional expansion I.

The 'Generative Dynamic Existence Unit on a Relational Basis', summarized in Table 1 above, can be used to design the specific appearance of a 'plane' moving toward a 'solid figure' through a qualitative transformation. This is to say that it can be used to conceptualize the appearance of the relationship between data and models unfolding on the premise of expanding dimensions in AI. This study proposes that data and as an algorithm, which can be represented within an upward dimensional coordinate space, function as:

○ Fragments and traces of human life (represented as data in Figure 1).

○ and algorithm-driven human purpose applied to the training data (represented as the learning algorithm generated from training data in Figure 1).

Within this framework, both data and algorithms can be regarded as entities in their own right. Consequently, the interrelationship between data and algorithms, and the developmental process that emerges from this interrelationship, can be understood not merely as fitting data to an algorithm or finding hidden patterns embedded in data but as an 'appearance of newly extended dimension (crSTR, represented as the 'model' in Figure 1).

This process can be accepted as the work of recreating the appearance of human life represented as data in an organized way that corresponds to another purpose of humans, namely the algorithm, while, in some cases, modifying even the purpose. Eventually, this structured representation of human life, which gets along with human purpose as embodied in the algorithm, progresses, as time goes on, increasingly towards a new existence, namely 'Another Bigger I Converging Newly in Days (ABICND)', which is constituted by the activation of hidden units generated by the model (López-Rubio 2021).

**E. Hypothesis Examination from Perspective of Systems Biology**

If we adopt an extended view of transforming a 'plane' into a 'solid figure'—not as a singular process but as a continuously progressing developmental trajectory,[8] then, its analogy with biological evolution, particularly in systems biology, becomes apparent. In biology, the stages of development proceed as follows: from DNA to proteins, from proteins to cells, from cells to tissues, and from tissues to organs (Noble 2006, 2012). This progressive biological development phase bears a striking resemblance to the scalable developmental phase in AI, where data and algorithms interrelate to shape a continuously evolving system.

In AI, the relationship between nodes is not a one-time occurrence but a series of continuous events

---

[8] Given that the performance resulting from the interrelation between data and algorithms assumes an open-world premise, this becomes a natural consequence. Furthermore, the increase in nodes within a DNNs can be seen as representing this phenomenon.

(Domingos 2015). Given that hidden units between nodes further underscore this dynamic nature, this process can be seen as analogous to biological scaling, where each developmental phase emerges within a relative space-time and subsequently leads to the formation of a higher-order system (Auffray et al. 2003). Just as in biological life phenomena, where successive phases emerge as part of a scalable development process, each stage in AI can be interpreted as progressively generating a higher-level representation through its evolving interrelations. If transforming this view into a holistic perspective, it becomes evident that there is a notable divergence from the conventional mechanical and linear perception of DNN. Furthermore, the prevailing approach to DNNs is an end-to-end methodology that indiscriminately converts all nodes into a numerical degree of gradient, constrained only by the limits of the mathematical tools employed (McCulloch and Pitts 1943). This indiscriminate conversion inevitably results in the distortion and deformation of the appearance created by the relationship between data and algorithms. Typical examples of unexpected problems caused by such distortion and deformation include overfitting, underfitting, and gradient vanishing (Bishop 2006; Flach 2012). It is also inevitable to observe hallucination (Bender et al. 2021; Sun et al. 2024), which is recently accepted as the most significant issue in the AI industry and academia, as a consequence of this process.

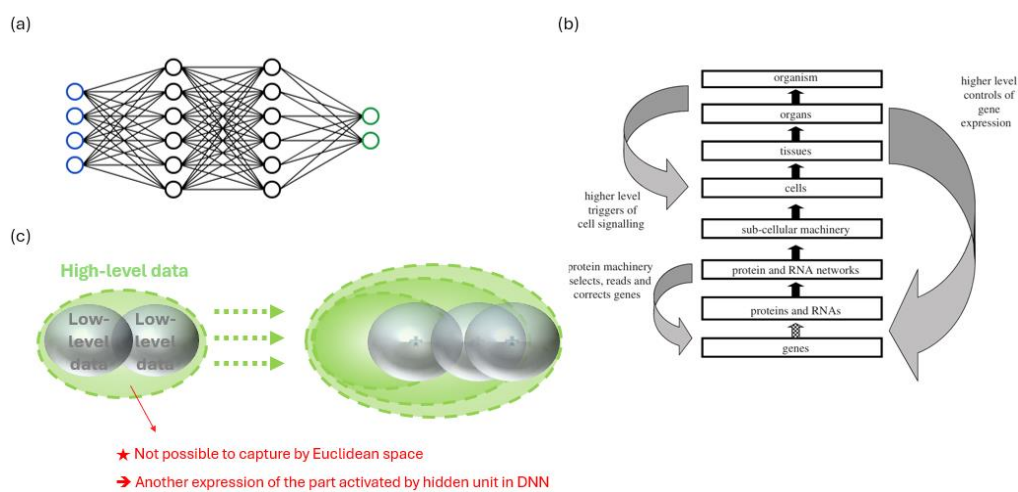

Figure 2. The relationship between nodes connected by complicated lines in (a), the causation represented by two-way in (b), and the green part in which crSTRs are manifested in (c) – which represent the common property causing the area of generation. (a) The conventional structure of DNNs. (b) Causation is, therefore, two-way, although this is not best represented by making each arrow two-way. A downward form of causation is not a simple reverse form of upward causation. It is better seen as completing a feedback circuit. (This image is adapted from (Noble 2012)) (c) The relationship between low- and high-level data features proceeding from a fixed perspective to an overall perspective (from one scale to another).

2) Second Conceptual Hypothesis – DCPSs of TC-EO

The initial conceptual hypothesis posits a process for capturing the expansion of dimensions accompanied by qualitative transformations, which differ from quantitative expansions in that they introduce fundamental structural changes rather than mere incremental scaling. While quantitative expansion simply increases the number or size of elements within a given framework, the qualitative transformations involve the emergence of new properties or relationships that redefine the framework itself. The second conceptual hypothesis, in contrast, seeks to elucidate the driving force that causes qualitative transformation as a concomitant feature of the expansion of dimensions that occurs in DNNs as data science. As has been demonstrated, the phenomenon of 'qualitative transformation' manifests in the process of creating performance as data and algorithms within the same dimension make relationships, whereby a relative space-time of another higher dimension is formed. Consequently, an additional perspective should be required to identify the catalyst that enables the qualitative expansion of dimensions in the process of creating crSTR and aiming for another entity, ABICND.

One hypothesis that aligns with this perspective is the DCPSs of TC-EO. The original DCPSs of TC-EO hypothesis is a philosophical concept related to the expansion of the subject derived from the core principle of 'Shendu (愼獨)' in the Confucian classic 'Zhongyong (中庸)'. Nevertheless, it is challenging to provide a comprehensive explanation of this philosophical concept in the context of this study.[9] Instead, this study posits that the functioning of the 'strange loop', as described in Hofstadter's Gödel, Escher, Bach: an Eternal Golden Braid (GEB), bears resemblance to the structure of DCPSs of TC-EO.[10] This study posits that the developmental process of DCPSs of TC-EO serves as the driving force that enables a system to go beyond its constraints and evolve into a higher-level system, allowing for the redefinition of its foundational properties and operational framework. Furthermore, this process can be applied to the expansion of dimensions that can occur in the interrelationships between data features, between data, and between data and algorithms.

### A. From 'Typographical Number Theory (TNT)' to (TNT+$G_\omega$)–PROOF–PAIR{a, a'}[11]

As stated in the preface of the 20th anniversary edition of GEB (Hofstadter 1999), Hofstadter attempts to address the question of how living beings can arise from non-living matter by shifting the focus from material components to abstract patterns. Among the abstract patterns, he places particular emphasis on the 'paradox of self-reference'. This self-reference paradox gives rise to a highly unusual loop, which begins with Russell and, in particular, is based on Gödel's discovery of 'Gödel numbering', and then progresses via the TNT string, 'I cannot be proved within the formal system TNT, that is, within the

---

[9] Consequently, I will also provide a more comprehensive introduction to this subject in my forthcoming paper, entitled "The Philosophical Basis of Data and Algorithm relationship Models".

[10] It is regrettable that this article does not have the space to present a detailed argument on this topic. Therefore, I will refer the reader to my other article on this subject, entitled "A Novel Approach to the relationships between Data Features – based on comprehensive examination of mathematical, technological and causal methodology."

[11] Hofstadter adapted the title of Whitehead and Russell's famous 1931 work, 'Mathematical Principles of Logic' (Whitehead and Russell 1927), to 'TNT' in order to make the paper more accessible to those unfamiliar with Gödel's proof. This study also follows Hofstadter in changing Russell's argument and proof to TNT for the sake of simplicity and clarity. This is because this study is not focused on proving complex formulas and mathematical rigor.

Principles of Mathematics (Whitehead and Russell 1927)'. Hofstadter does not view this loop as merely the 'essential incompleteness' of Gödel's formalization of mathematics. Instead, he aims to examine the generality of proofs by introducing $G$ to reinforce self-reference (Hofstadter 1999).

$$(TNT+G_\omega)\text{–PROOF–PAIR}\{a, a'\} \qquad (1)$$

Furthermore, the question of why $G_{\omega+1}$ cannot belong to the axioms created by $G_\omega$ can also be raised in the new formula compiled by Hofstadter. Hofstadter posits that this difficulty arises from the fact that the properties of the system are reflected in the concept of proof pairs, which are then used against the system in a manner analogous to Gödel's trick. Once a system has been clearly defined and placed within a conceptual framework, it becomes susceptible to critique. In summary, once the capacity for self-reference is established, the system is susceptible to a vulnerability inherent in its own structure. This vulnerability arises from the fact that, once all the elements, $G$s, are integrated into TNT in a well-defined manner, there must be another, unforeseen $G$ that cannot be captured in the axiomatic scheme. (Hofstadter 1999).

This contradiction can only be interpreted as self-destructive incompleteness if it is considered from the perspective of the system in isolation, distinguishing it from general incompleteness, which may not necessarily undermine the system's integrity but instead reveal its inherent limitations and scope. Even when viewed through the lens of completeness, the infinite expansion of TNT as postulated by Hofstadter's new formula ultimately leads to the same conclusion: that it is self-destructive incompleteness. This is due to the fact that incompleteness is an intrinsic aspect of TNT (Hofstadter 1999). As previously stated, this is an unavoidable outcome when contemplating the essence of mathematics, which establishes its own axioms in pursuit of uniqueness and completeness on the assumption of a closed world (Whitehead and Russell 1927).

### B. "(TNT+$G_\omega$)–PROOF–PAIR{a, a'}," "Expansion of Dimensions as an Open World," and "DCPSs of TC-EO"

This study interprets the practical significance of Hofstadter's new formula—primarily based on the repeated application of Gödel's argument—as the incorporation of the idea of "Expansion of Dimensions as an Open World" into the existing Gödelian framework. By doing so, the assessment of completeness should not be limited to a closed world evaluated solely on the basis of the lower-dimensional framework. Instead, even if it cannot be fully defined, an approach to dynamically engaging with the expanding world based on higher-dimensional perspectives as an open world becomes feasible. In this regard, this study suggests that "Expansion of Dimensions as an Open World" corresponds to the "Expansion of Dimensions accompanied by Qualitative Transformation" in the CFP theory discussed earlier. Furthermore, it proposes that the hypothesis enabling such "Expansion of Dimensions accompanied by Qualitative Transformation" is the DCPSs of TC-EO. The conceptual hypothesis of the DCPSs of TC-EO focuses on the dynamic expansion of the world as an open system, irrespective of whether the axiomatic system is complete or not. This approach allows for the evolution of relationships between Paradox and Stratification, going beyond the Duplexity-Contradiction relationship found in Hofstadter's new formula. This conceptualization facilitates the transformation from the Duplexity-Contradiction relationship observed in Hofstadter's new formula to the Paradoxical and Stratified structure. In other words, for the Duplexity-Contradiction relationship to evolve into a Paradox-Stratified relationship, a fundamental shift in worldview is first required—one that is achievable within DNNs as a data science framework grounded in real-world complexity rather than within the closed mathematical axiomatic world.

If this is the case, it becomes crucial to examine the specific characteristics inherent in data science as an open system that enables the transformation from the Duplexity-Contradiction relationship to the Paradox-Stratification relationship.

As previously discussed, Gödel's proof relies on the mathematical axiomatic world, which fundamentally operates on recursion to determine the completeness or incompleteness of an original system. In contrast, data science as an open system does not prioritize static completeness but instead centers on the continuous creation and transformation of relationships among data. In data science, data is not simply an isolated entity but a dynamic trace of real-world interactions, and its relationships with other data continuously shape and redefine its structure. Therefore, data science is primarily evaluated based on its capacity to generate dynamism rather than static completeness. To activate the generation of dynamism, the recursive process based on identity is applied through a fitting process that structures relationships not only between features within data but also between data points themselves, as well as between data and algorithms. In this way, the structuring of data relationships bears similarities to recursive processes.

This "structuring of relationships in a manner similar to recursion" closely parallels the "process of achieving paradoxical synergetic effects of economic agents" as proposed in the Convergent Paradigm theory (CP theory) within the field of social science. In the formation and maintenance of economic and social relationships, the maximization of an economic entity's interests is not necessarily determined by direct transactional calculations (i.e., a simple give-and-take arithmetic model). Instead, the pursuit of interests can also be achieved through cooperative relationships, whereby an entity concedes or sacrifices short-term benefits in favor of long-term synergy. This dynamic often entails an entity identifying with the interests of others, fostering a broader, collective benefit. This method of achieving unexpected benefits through recursive, cooperative engagement can be likened to the creation of community bonds—expressed conceptually as crSTR or ABICND. Although this approach may appear to entail short-term losses from the perspective of individual economic agents, the long-term synergistic effects generated by community-based ties often compensate for and surpass these losses. Thus, the actions of agents based on the relationships with each other (understood as 'Duplexity'), initially perceived as Contradictory, ultimately yield Paradoxical outcomes that enable the emergence of a more complex and higher-order entity (understood as 'Stratification')—eventually understood as crSTR or ABINCD as a dimensional system of relationships.

In short, the field of data science, by virtue of its open-world characteristics, structures data through recursive-like relational processes even when apparent identicality between data features, data points, and algorithms is not readily found out. This dynamic structuring approach extends beyond mere quantitative expansion and instead facilitates "Expansion of Dimensions accompanied by Qualitative Transformation." Within this transformation, the transition from the Duplexity-Contradiction relationship to the Paradox-Stratification relationship reflects a higher-order process similar to the "paradoxical synergetic effect of economic agents" previously described.

To apply Hofstadter's formula (Hofstadter 1999), which relies on the repeated application of Gödel's theorem, within the context of data science, it is necessary to establish "Duplex-Contradictory-Paradoxical-Stratified structures (DCPSs)" as a distinct framework. This methodology encapsulates the recursive-like binding characteristics necessary for structuring data relationships within an open-world paradigm. Unlike conventional mathematical representation within Euclidean space, the evolution from "Duplexity-Contradiction" to "Paradox-Stratification" structures requires a transformative process beyond standard formal logic.

Meanwhile, for this evolution to occur, it must be accompanied by the "Thorough Closure - Eternal Opening (TC-EO)" process, wherein lower-dimensional entities strive for higher-dimensional openness through an initial phase of closure. This transformation does not unfold as an overt process but rather as an implicit, time-dependent phenomenon. In the context of Gödel's argument, Thorough Closure (TC) is manifested as systemic incompleteness, while Eternal Opening (EO) contradicts the principles of a closed mathematical system. However, in an open-world paradigm, these contradictions can serve as a generative force, facilitating the emergence of higher-order structures. Crucially, this process remains concealed, revealing itself only a posteriori as its consequences unfolds over time. For this reason, this study evaluates the development of DCPSs of TC-EO as a systematic means of applying Gödelization to individual cases, in alignment with Hofstadter's framework (Hofstadter 1999).

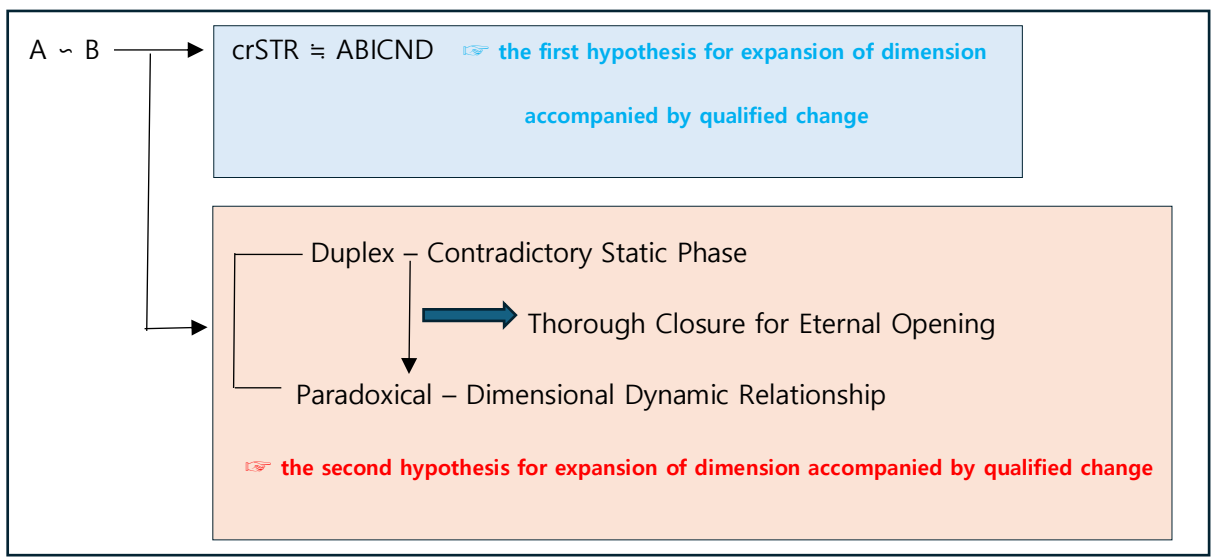

Table 2. New Conceptual hypothesis for dimensional expansion II

**3) Summary – CFP theory functioned as new data science**

This study has thus far proposed two conceptual hypotheses to represent the 'dimensional expansion accompanied by qualitative transformation' as a new developmental phase created by the unique 'relationship modeling between data and algorithms' in DNNs. This approach builds upon the achievements of existing physics and biology, along with Hofstadter's reinterpretation of Gödel's proof, formalized as (TNT+$G_\omega$)–PROOF–PAIR{a, a'}. These two conceptual hypotheses offer new insights into the relationships and structural integration between data features, between data, and between data and algorithms, ultimately providing a systematic framework for the convergence and fusion of high-dimensional and low-dimensional data structures. Accordingly, this study introduces and names this theoretical framework the Convergent Fusion Paradigm (CFP) theory, positioning it as a new geometric perspective within data science.

## 3. Application of two conceptual hypotheses featuring CFP theory to Riemannian manifold

The next step is to apply CFP theory to existing data generation models. CFP theory enables a precise understanding of the data generation process and offers novel insights by providing an alternative interpretation of the Riemannian metric, which is frequently employed in the data manifold hypothesis.

**1) Data manifold hypothesis and generative models**

The concept of a 'manifold' originates from mathematics and describes a space that, while locally resembling Euclidean space, may have a much more complex global structure (Kelley 2017). The data manifold hypothesis applies this concept to data, proposing that most naturally occurring high-dimensional datasets actually reside near low-dimensional non-linear manifolds (Belkin and Niyogi 2003; Dominguez-Olmedo et al. 2023; Hastie and Stuetzle 1989; Smola et al. 2001). This hypothesis plays a crucial role[12] in machine learning and data analysis, serving as a foundation for understanding and utilizing the structural properties of data.

---

[12] The geometric structure of actual manifolds not only plays a crucial role in various machine learning tasks such as classification, clustering, density estimation, representation learning, and transfer learning, but also finds applications in diverse fields such as computer vision (Arvanitidis et al. 2018, 2021; Tosi et al. 2014), robotics (Beik-Mohammadi et al. 2021, 2023; Scannell et al. 2021), human motion capture (Tosi et al. 2014), and protein sequence analysis (Detlefsen et al. 2022). These applications demonstrate the strong inductive bias that manifolds provide (Dominguez-Olmedo et al. 2023).

Generative models, on the other hand, provide an appealing framework to approximately learn the data manifold through differential geometric studies (Arvanitidis et al. 2018). Notably, equipping a generative model with a Riemannian metric (Arvanitidis et al. 2018, 2019, 2020, 2021; Tosi et al. 2014) has presented a more effective approach to learning manifolds while addressing the inherent identifiability problem in existing generative models. When the two conceptual hypotheses of CFP theory are integrated with the Riemannian manifold hypothesis, they enable a more organic representation of the structured, embedded low-dimensional non-linear manifold within the high-dimensional feature space. This results in a more precise capture of the data generation process. As previously discussed, when dimensional expansion is approached by means of conventional mathematical methods within Euclidean space, it remains confined to a quantitative framework. However, CFP theory, grounded in its two conceptual hypotheses, facilitates a form of dimensional expansion that incorporates qualitative transformations. Through this process, low- and high-dimensional data structures form a convergent, integrative structural relationship. In other words, CFP theory offers a geometric framework for the convergence and fusion of low- and high-dimensional data as a sort of relationship that is difficult to articulate within Euclidean space. This framework, in turn, provides new insights into the data generation process. To further explore this, we first examine how the existing Riemannian metric can be theoretically understood from the perspective of CFP theory.

## 2) Expressivity of matrices and vector inner products in Euclidean coordinate space

Before delving into this topic, two key mathematical aspects should be addressed: the mathematical properties of matrices and the vector inner product, which serves as the foundation for defining the Riemannian metric.

The initial topic of discussion will be the representation of matrices in Euclidean space. To illustrate, in the context of a two-dimensional coordinate plane, a matrix is essentially represented as $m \times n$, where $m$ and $n$ are positive integers. In order to represent a matrix that exceeds this in a two-dimensional Euclidean space, it is necessary to make quantitative adjustments to each row and column. However, attempting to explain all matrices as an extension of the dimension would exceed the existing two-dimensional Euclidean space. In such cases, the expanded dimension is expressed as a diagonal matrix. Consequently, since Euclidean spaces is linear, the expansion of the dimension is expressed as quantitatively expanded plane.

Next, the geometric interpretation of the vector inner product in coordinate space will be examined. It can be understood as a process wherein one vector becomes incorporated into another. From the perspective of CFP theory, this corresponds to a scenario in which object A converges with object B and merges into a higher dimension, a phenomenon described as a 'higher-dimensional extension.' However, a distinction exists between the vector inner product relationship and the relationship facilitating the generation of crSTR in CFP theory. While the vector inner product is constrained by its reliance on a linear coordinate space, requiring that one of the entities $A$ or $B$ be the basis for inclusion, CFP relationships are not bound by such constraints. Consequently, there is no necessity for one entity to be defined as the basis for inclusion.

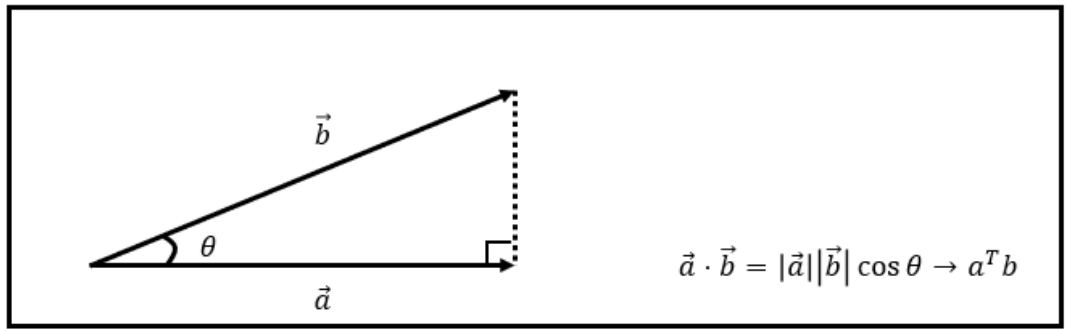

Figure 3. The schematic mechanism of the vector inner product incorporating the one into the other.

**3) Examining Ontological Meaning of Riemannian Metric**

When understood in this way, the Riemannian metric $\dot{\gamma}^T M_\gamma \dot{\gamma}$ can be interpreted as assigning each matrix element of $M_\gamma$ (the Riemannian matrix, or more precisely, a symmetric positive definite matrix) as an inner product for each dimension. In this case, $\gamma(t)$ is a smooth curve, $\dot{\gamma}(t)$ is the velocity of the curve, and $M_\gamma$ is a symmetric positive definite matrix that acts akin to a local Riemannian metric.

From this perspective, the fundamental unit of the Riemannian metric can be understood ontologically as follows: it represents the transformation of a "static, isolated, formal existence" ($\gamma(t)$) into a "dynamic, relational, and substantial existence" ($\dot{\gamma}^T M_\gamma \dot{\gamma}$). This transformation occurs by determining the interrelation between the self ($\gamma$) and its recursive counterpart ($\gamma^T$, the transposed matrix). The dynamic existence represented by this transformation not only manifests its magnitude on the coordinate plane as an inner product but also signifies an extension into a higher-dimensional framework.

To grasp the necessity of using a transposed matrix in dimensional changes and interpreting it as an interrelation between the self ($\gamma$) and its recursive counterpart ($\gamma^T$), an ontological understanding is required. In an isolated dimension, the concept of existence can remain static. However, when dimensions change, a static notion of existence becomes insufficient, as any transformation in dimension inherently involves a relational engagement with other entities, including the self.

Consequently, dimensional transformation necessitates shifting from a static, isolated notion of existence to a dynamic, relational one. Reversing this understanding, it is only through the establishment of these dynamic, relational concepts that a dimensional change involving a manifold-orienting concept rather than a plane-based one is possible. To establish this dynamic and relational concept, the mathematical formulation of a Riemannian metric places the Riemannian matrix (typically the identity matrix in lower dimensions, transitioning to a higher-dimensional matrix in manifold extensions) between the self ($\gamma$) and its recursive counterpart ($\gamma^T$), positioning it as a structural intermediary that facilitates higher-dimensional existence, which corresponds to the position of crSTR, orienting towards a higher-dimensional existence, in terms of CFP theory.

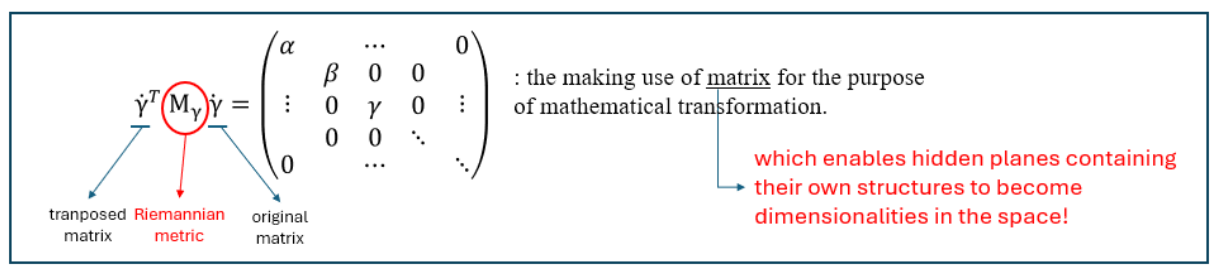

Figure 4. Riemannian metric and diagonalized matrix.

### 4) Understanding from Perspective of CFP Theory – Possibilities and Limitations

From a philosophical and ontological perspective, if the Riemannian metric is understood as a mathematical concept that transforms a mere static and isolated unit of existence on a coordinate plane into a dynamic relational unit accompanied by dimensional qualitative transformation, then within CFP theory, the Riemannian metric itself can be interpreted as crSTR. According to CFP theory, the entities *A* and *B* should not be regarded as static, isolated existences but inherently possess relational meaning. Therefore, the entity as a fundamental unit of existence will transform into ABICND through crSTR.

Furthermore, the notion that a unit of existence is established through crSTR implies that it should undergo a process of convergence and fusion with its recursive counterpart. This necessitates passing through DCPSs of TC-EO, which constitutes the second conceptual hypothesis of CFP theory. While the relationship of "Duplexity-Contradiction" can be represented within Euclidean space, the dynamic process wherein this relationship undergoes thorough closure, leading to paradoxical eternal opening and the creation of new unexpected higher-dimensional entities, cannot be fully captured in the space. Additionally, due to the geometric modular restriction that Euclidean space should be differentiable, the concept of "Duplex-Contradiction" in CFP theory is reduced to the "reflexive manifestation of an ontological unit." In Euclidean space, "Duplex-Contradiction" manifests merely as "reflexive manifestation through the process of contrast." This reduction in the Duplex-Contradiction (DC) concept inevitably diminishes the associated generation-related Paradox-Stratified (PS) concept. Owing to the inherent linearity of Euclidean space, dimensional transformation regresses into a linear expression of mere "quantitative expansion" rather than "generation accompanied by qualitative transformation," resulting in a sort of 'retardation-shrinkage'. Consequently, the TC-EO process, which facilitates the DCPS transformation, is either omitted or distorted.

A significant point to note is that the ontological regression-shrinkage of the Riemannian metric within the framework of CFP theory, constrained by Euclidean space, aligns with the contradiction and incompleteness issues inherent in Hofstadter's formula, which is based on the repeated application of Gödel's theorem. To effectively address this issue within data science, the regression-shrinkage problem of the ontological meaning of the Riemannian metric should be resolved, just as the separate establishment of DCPSs of TC-EO was required alongside Hofstadter's formula. This necessity has been recognized in recent research (Arvanitidis et al. 2020). The core of this approach (Arvanitidis et al. 2020) involves utilizing the Jacobian inverse function in the formulation of the Riemannian metric. Therefore, to properly understand this, we will examine the studies (Arvanitidis et al. 2018, 2020, 2021) that have investigated the appropriate formulation of the Riemannian metric in VAEs structural spaces, particularly in the ambient (input) space and latent space.

## III. Examination of Endowing Inner Structures of VAEs with Riemannian Metrics.

### 1. Preliminary Work – Understanding Spatial Structure Inside VAEs

Before summarizing and analyzing Arvanitidis's research pertaining to the establishment of the Riemannian metric, it is crucial to develop an understanding of the spatial structure within VAEs. This is because a thorough comprehension of the Riemannian metric, beyond its mathematical formulation, requires an understanding of these geometric spatial structures.

To grasp the overall structure, two figures have been provided: a flat representation of the VAEs viewed from above and a three-dimensional representation viewed from the side. Based on this, we examine how the structures of the ambient (input) space, latent space, and ambient (output) space within VAEs can be categorized according to their magnitude and path perspectives. This analysis is necessary because, while these spaces may appear structurally divided, they actually overlap.[13] Understanding this superposition is essential to accurately conceptualizing the processes within VAEs. A correct interpretation of these overlapping spaces allows us to extend and measure the Euclidean metric, originally set in the ambient (input) space, into the latent space. Furthermore, it enables us to differentiate between the compression process from the ambient (input) space to the latent space and the generation process from the latent space to the ambient (output) space.[14] This differentiation provides valuable insights into the mechanisms of compression and generation in VAEs.

---

[13] In conventional physical sciences, the distinction between superposition and overlap is not explicitly defined. However, this paper aims to clarify the difference between these two terms.

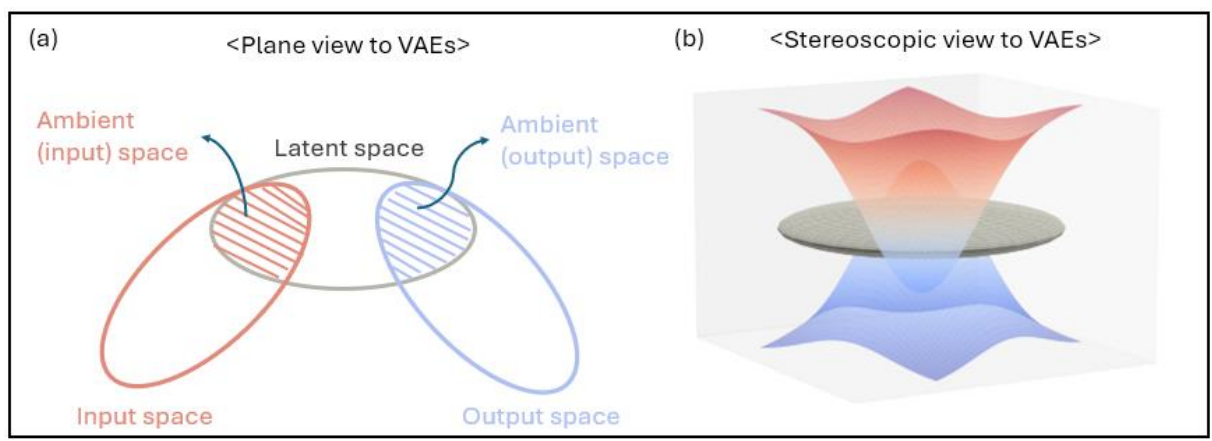

Figure 5. Two viewpoints on the structure of VAEs. (a) A schematic representation of the VAEs structure viewed in a planar perspective when laid flat. (b) A schematic representation of the VAEs structure viewed from the front in a stereoscopic perspective. The red-shaded area represents the ambient (input) space, the blue-shaded area represents the ambient (output) space, and the grey elliptical area represents the latent space.

The structures of the ambient (input) space, latent space, and ambient (output) space within VAEs can be classified according to their respective magnitude and path perspectives, as follows:

○ *Ambient (Input) Space*: On a magnitude unit, the ambient (input) space embraces high-dimensional information and should therefore be understood by means of a macroscopic perspective. However, since each data point within this high-dimensional space can exhibit its own distribution, the space can also be analyzed by means of a microscopic perspective when adopting a linear path unit.

○ *Latent Space*: Although it is a lower-dimensional space than the ambient space, the latent space is still a manifold embracing its own dimensional structure. Therefore, on a magnitude unit, it can be seen by means of a microscopic perspective due to its lower dimensionality relative to the ambient space, but also by means of a macroscopic perspective because it encompasses its own internal dimensions.

○○ Meanwhile, since the latent space has a lower dimensionality than the ambient space, it can be easily analyzed on a path unit. If each data distribution is examined along a linear path, the latent space can be interpreted by means of a microscopic perspective. Notably, this method of understanding the latent space overlaps with the microscopic perspective of analyzing the ambient (input) space. This overlap suggests that, when viewed microscopically, the data density distribution in the ambient space can be pulled back into the latent space.

---

[14] Meanwhile, the overlap assumed in the compression process from the ambient (input) space to the latent space differs from the superposition that is based on the generation process. The former overlap can be measured by converging to Euclidean space, whereas the latter cannot be accurately measured in the strict sense since it must go beyond Euclidean space. This paper proposes the use of Hilbert space to make this measurement possible.

○ ***Ambient (Output) Space***: Since the ambient (output) space represents the reconstructed and generated data, it does not require a microscopic perspective, regardless of a path or magnitude unit. Instead, it should be examined by means of a macroscopic perspective, as it encapsulates the high-dimensional process of reconstruction and generation. Notably, the macroscopic perspectives of the latent space and the ambient (output) space can overlap in terms of reconstruction and generation. This overlap is understandable given that the VAEs generator operates based on probability distribution principles.

Finally, it is crucial to compare the overlap between the ambient (input) space and latent space, which is derived from a microscopic perspective, with the overlap between the latent space and ambient (output) space, which is derived from a macroscopic perspective. The key distinction between these two forms of overlap is that the former involves extending the Euclidean metric from the ambient (input) space into the latent space, making it measurable, while the latter does not possess this property, making Euclidean-based measurement impossible. This paper argues that this difference indirectly indicates the existence of a hidden space where the generation process manifests. Due to this distinction, the latter overlap should be interpreted as superposition, distinct from the former. Consequently, this paper suggests that approximating the generation process more accurately requires an alternative space beyond conventional Euclidean space. However, there is currently no research explicitly addressing this difference in the nature of these overlapping structures.

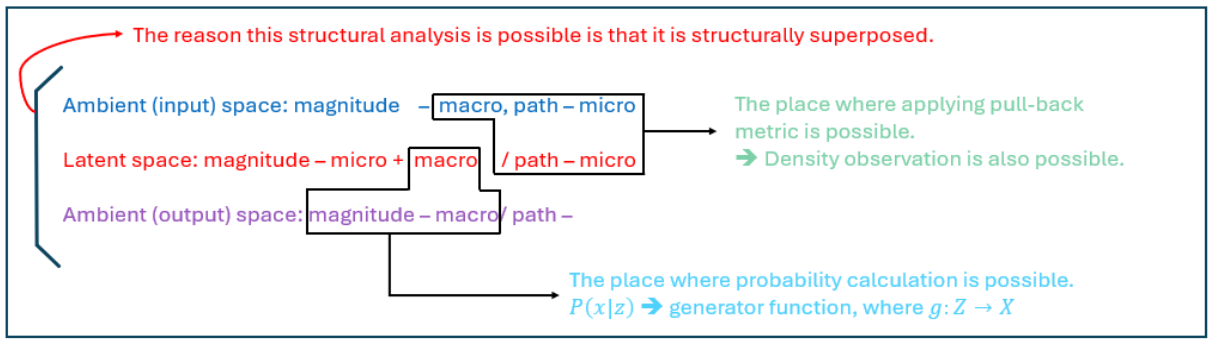

Figure 6. Interpretation of structured VAEs.

## 2. Analysis of (Arvanitidis et al. 2018)

**1) Structurization method – entitling latent Space with Riemannian metric**

First, it is necessary to pay attention to the way of structuring, when it comes to modelling, the ambient space $\chi$ and latent space $Z$ in the previous studies that appeared before (Arvanitidis et al. 2018). [Previous researches have shown that] generative models such as VAEs and generative adversarial networks (GANs) provide a flexible and efficient parametrization of the density of observations in an ambient space X through a typically lower dimensional latent space $Z$ (Goodfellow et al. 2014; Kingma and Welling 2013; Rezende et al. 2014). [However,] these generative models are subject to identifiability problems, such that different representations can give rise to identical densities (Bishop, 2006). This implies that straight lines in Z are not shortest paths in any meaningful sense, and therefore do not constitute natural interpolants (Arvanitidis et al. 2020, p1).

To overcome this issue, it has been proposed to endow the latent space with a Riemannian metric such that curve lengths are measured in the ambient observation space $\chi$ (Arvanitidis et al. 2018; Tosi et al. 2014). This approach immediately solves the identifiability problem. In other words, this ensures that any smooth invertible transformation of $Z$ does not change the distance between a pair of points, as long as the ambient path in $\chi$ remains the same (Arvanitidis et al. 2018, p1).

This can be expressed in mathematical terms as follows,

$$\text{Length}[f(\gamma_t)] = \int_0^1 \|\dot{f}(\gamma_t)\| dt = \int_0^1 \|J_{\gamma_t}\dot{\gamma}_t\| dt, \quad J_{\gamma_t} = \frac{\partial f}{\partial z}|_{z=\gamma_t} \quad (2)$$

$$\|J_{\gamma_t}\dot{\gamma}_t\| = \sqrt{(J_\gamma\dot{\gamma})^T(J_\gamma\dot{\gamma})} = \sqrt{\dot{\gamma}^T(J_\gamma^T J_\gamma)\dot{\gamma}} = \sqrt{\dot{\gamma}^T M_\gamma \dot{\gamma}} \quad (3)$$

Here, $J_\gamma \dot{\gamma}$ is the Jacobian matrix of $\dot{f}(\gamma_t)$). Furthermore, in Eq. (2), $M_\gamma = J_\gamma^T J_\gamma$ is a symmetric positive definite matrix that defines the Riemannian metric[15].

To summarize, (Arvanitidis et al. 2018) defines the Riemannian metric using the Jacobian $J_\gamma$ in the latent space based on the deterministic case and allows the measurement of the curve lengths in the ambient space through this. This method can be extended to the stochastic case.

---

[15] In the context of Riemannian geometry, the Jacobian matrix $J_\gamma$ is always positioned on the left and right of the Riemannian metric, in the form of a transpose matrix and an original matrix, respectively. Additionally, the Riemannian metric placed between them represents a level that is one higher than the corresponding dimension. This implies that the geometric structure of this Riemannian metric, when represented in terms of the coordinate plane, should have the form of a point, which symbolizes a virtual 'linear (higher dimension)', enabling a sort of embedded form to contain 'lower dimensions'. From the perspective of the CFP theory, this can be seen as a form in which existence *A* converges and merges with the recursive form *A'* into a created relative Space-Time in Relationships (crSTR), because crSTR must represent a higher dimension than the existences of *A* and *A'*. The application of CFP theory will be detailed in the latter part of the paper.

## 2) Evaluation

However, the method proposed by (Arvanitidis et al. 2018) –setting the Riemannian metric in the latent space by means of Jacobian $J_\gamma$ and measuring it in the ambient space through the Euclidean measure – raises some concerns. As shown in Eq. (3), the norm of $\|J_{\gamma_t}\dot\gamma_t\|$ is merely a single point in the ambient space. Consequently, for the Euclidean measure to be applicable, the unit length should be represented as an integral between 0 and 1 in Eq. (2). This process ensures the feasibility of measuring curve lengths in the ambient space $\chi$.

However, complications arise when extending this method to the stochastic case. In the stochastic case, the measured distance in the ambient (input) space should be pulled back into the latent space (requiring a type of pull-back process), yet the precise explanation for this step remains unclear. It is important to note that the ability to measure in the ambient (input) space does not inherently equate to the ability to transfer those measurements back into the latent space. Acknowledging this issue, the subsequent study (Arvanitidis et al. 2020) introduces the inverse function of the Jacobian, allowing for the metric set in the ambient (input) space and the measured results to be transferred together into the latent space. Within this context, whether the approach of (Arvanitidis et al. 2018) can be strictly considered as a pull-back metric or not remains questionable.

(Arvanitidis et al. 2020) appears, nevertheless, to interpret the method proposed by (Arvanitidis et al. 2018) as a pull-back metric. This is due to the fact that, in the stochastic case, the Jacobian matrix becomes $J_{g_\epsilon}(z)$ and the Riemann metric is also represented as $M_\chi(\mu(z) + \text{diag}(\epsilon) \cdot \sigma(z))$, if observing the overlap between the ambient (input) space and the latent space, from the perspective of generator. However, without setting the ambient metric in the ambient space, it is impossible to transfer the measurement results into the latent space. Considering this, (Arvanitidis et al. 2018) appears to have measured data density observations in the latent space using the Radial Basis Function (RBF) neural network. While observing data density in the latent space is possible and furthermore, it is not meaningless in that it supplements data distribution by means of the mean and variance of the Gaussian distribution, measuring only in the latent space presents a significant limitation—it cannot fully reflect the density and curvature of data present in high-dimensional space.

## 3. Analysis of (Arvanitidis et al. 2020)

### 1) Structurization Method

## A. Endowing Ambient Space with Riemannian Metric and Pulling it Back into Latent Space (So-called Pull-back Metric)

As previously stated, (Arvanitidis et al. 2020) sets the Riemannian metric in the ambient space and allows this metric and its measured results to be pulled back into the latent space (so-called pull-back metric).

To achieve this, (Arvanitidis et al. 2020, p2) assumes that ambient space $\chi$ is equipped with the Euclidean metric $M_\chi(x) = \mathbb{I}_D$ and its restriction is utilized as the Riemannian metric on $T_\chi \mathcal{M}$, which indicates tangential mapping of manifold $\mathcal{M}$. Since the choice of $M_\chi(\cdot)$ has a direct impact on manifold $\mathcal{M}$, we can utilize other metrics designed to encode high-level information.

In light of the aforementioned, the pull-back metric presented in (Arvanitidis et al. 2020) is as follows.

$$M_\chi\big(\mu(z) + \text{diag}(\epsilon) \cdot \sigma(z)\big), \quad g(z) = \mu(z) + \text{diag}(\epsilon) \cdot \sigma(z) \tag{4}$$

Based on the above, the conditions for establishing the pull-back metric are as follows:

(a) A Riemannian metric ($M_\chi$) must be defined in the ambient space.

(b) The Riemannian metric ($M_\chi$) must be pulled back into the latent space through the inverse Jacobian $J_{\phi^{-1}}$.

$$M(z) = J_{\phi^{-1}}(z) M_\chi\big(\phi^{-1}(z)\big) J_{\phi^{-1}}(z) = M_z = J_z^T M_\chi\big(g(z)\big) J_z \tag{5}$$

Through the processes (a) and (b), the metric information of the ambient space is expressed in the latent space, making it measurable within the latent space, as indicated in Eq. (5). In other words, the distortion and curvature of the ambient space can be reflected in the measurement of the curve length in the latent space through this process.

## B. Interpretation from Perspective of CFP theory

To understand the structural characteristics created by the pull-back metric outlined above, it is first necessary to conduct a logical and philosophical analysis of the implications of using the inverse function of the Jacobian. This logical and philosophical analysis should be conducted from the perspective of CFP theory.

As discussed earlier, (Arvanitidis et al. 2018) does not explicitly mention the inverse function of the Jacobian in the deterministic case. However, (Arvanitidis et al. 2020) introduces the inverse function of the Jacobian, making it possible to incorporate a form of reversed time within the Riemannian metric. This shifts the conventional Euclidean understanding of space, which is limited to quantitative expansion (including compression and restoration), towards dimension expansion accompanied by qualitative transformation. This transformation is feasible due to the unique structural characteristics of the spatial configuration within VAEs, as previously described.

To further elaborate, the reason why the inverse function of the Jacobian can actually be implemented in the structure of VAEs is that the ambient (input) space and the ambient (output) space overlap structurally, even if their orientation and position are reversed. In this case, the notion of overlap should not only be understood as spatial within VAEs but also as temporal, given the nature of the inverse function. In other words, this does not imply that the ambient (input) space is reversed and attached to the ambient (output) space. Instead, it suggests that the procedures typically performed later in the ambient (output) space can be executed first within the ambient (input) space—essentially bringing the future into the present. Consequently, this process enables to embed the relative time reversibility within the Riemannian metric, which only has a spatial understanding of the dimensional expansion (compression, reconstruction). In short, the function of the Jacobian makes the reversed relative time operate within the Riemannian metric. This reversed time insertion extends beyond quantitative dimensional expansion (compression, reconstruction) by enabling qualitative transformation-based dimensional expansion (including generation) within the existing Riemannian metric framework. In this sense, by employing the inverse function of the Jacobian, the relationship between high-dimensional and low-dimensional data can be organically established rather than merely approximating the relationship between two in a linear manner.

In summary, using the inverse function of the Jacobian means inserting reversed time into the Riemannian metric, opening the door of a new possibility, 'qualitative transformation of relative space-time' to the existing Euclidean interpretation of dimensions limited to 'quantitative expansion of space'. This advanced understanding of dimensions not only allows for the compression and reconstruction of data but also facilitates an understanding of data generation within relative space-time. More significantly, this framework enables the organic processing of low- and high-dimensional information within the data.

From a deeper CFP theory perspective, the Riemannian metric serves as a form of manifold hypothesis, allowing for a "reflexiveness-contractiveness" structure through the use of transposed and original matrices in the Euclidean coordinate plane. This results in a framework that closely resembles duplexity-contradiction, a core concept of CFP theory.

Through this framework, mechanisms such as compression, approximation, and restoration can be mathematically formulated based on the traditional Euclidean understanding of space as a quantitative expansion mechanism. However, generation as a concept is fundamentally different and cannot be captured through compression, approximation, and restoration alone. The CFP theory proposes that to account for generation, the structure must extend beyond "reflexiveness-contractiveness" into a "Duplexity-Contradiction-Paradox-Stratification (DCPSs)" structure (see Section II. 2).

At this point, multi-dimensionality is no longer a simple quantitative extension but rather a new dimensional creation accompanied by qualitative transformation. To achieve this, a temporal element must be embedded within the spatial structure as a relational component. CFP theory describes this as "creating the relative Space-Time in Relationship (crSTR)."

To sum up, using the inverse function of the Jacobian introduces the reversed time into the Riemannian metric, transforming the Riemannian manifold hypothesis from merely expressing quantitative dimensional expansion to capturing qualitative transition of relative space-time. This transformation shifts the focus beyond compression-restoration toward generation, forming the basis for what can be called the "Generative Hypothesis of Riemannian Manifold." The CFP theory's DCPSs mechanistic structure enables this generative hypothesis through crSTR, thereby providing a framework that captures both dimensional expansion and data generation in high-dimensional structures that manifests in ambient (output) space.

**C. Re-interpretation of Macroscopic View on Posterior Distribution via Lens of CFP Theory**

The CFP theory summarized in the previous section provides a framework for reinterpreting the generation process of the generator, which is based on the existing stochastic approach that operates in both the latent space and the ambient (output) space within VAEs. These two spaces overlap in the macroscopic view when considering magnitude units.

As is well known, the outcome of a generator can be expressed as a conditional probability distribution, $p(x|z)$. The nonlinear relationship of $p(x|z)$ from $p(z)$ can be estimated using a DNN, which functions as a Universal Approximator (UA). To achieve this, first, assuming the prior distribution $p(z)$ as a Gaussian distribution, the conditional probability $p(x|z)$ can also be modeled as a Gaussian distribution, where a sample $z$ drawn from $p(z)$ is input into the neural network.

$$p(x|z) = \mathcal{N}\big(x|\mu(z), \mathbb{I}_D \cdot \sigma^2(z)\big), \quad g(z) = \mu(z) + \text{diag}(\epsilon) \cdot \sigma(z) \tag{6}$$

where $\mathcal{N}$ is normal distribution, $\mathbb{I}_D$ is identity matrix with d-dimension, $x$ is ambient output variable, $z$

is latent variable. $\mu(z)$, the mean value and $\sigma^2(z)$, the variance are the function with respect to latent variable $z$.

However, if the latent space is represented solely as a prior distribution, it does not reflect the characteristics of the actual data $x$ at all. As previously mentioned, $p(z)$ is simply assumed to be a Gaussian distribution, without incorporating any features of the real dataset. To accurately generate data that preserves the characteristics of the original dataset, the latent space should be in a well-differentiated state according to those characteristics.

In so doing, this requires examining the overlapping internal spatial structures of VAEs. As discussed earlier, the ambient (input) space and the ambient (output) space overlap as a single ambient space, mediated by the latent space. Consequently, the posterior distribution $p(z|x)$ from the ambient (input) space can serve as a substitute for $p(z)$. However, the posterior distribution $p(z|x)$ is typically highly complex and unknown, making it impossible to forgo the use of the prior distribution $p(z)$. As highlighted in (Arvanitidis et al. 2020), this issue leads to arbitrary extrapolation, to so-called regions of latent space without no latent codes. This poses a significant impediment to the generator's performance.

However, utilizing the pull-back metric to bring the Riemannian metric set in the ambient (input) space to the latent space, can have an effect similar to employing the posterior distribution, as it effectively encodes high-dimensional information from the dataset. Nevertheless, the exact reason why importing the Riemannian metric from the ambient (input) space to the latent space via the pull-back metric yields a similar effect to using the posterior distribution—referred to as *the quasi effect to the posterior distribution*—remains unclear. This is where the geometric structure provided by CFP theory can offer an explanation.

To better understand this, we reorganize the abovementioned probability distribution discussions by means of a macroscopic perspective on a magnitude unit, in relation to the internal structural space of VAEs. Under this framework, the conditional probability distribution $p(z|x)$ is based on the ambient (input) space, the posterior distribution $p(x|z)$, as the generator's outcome is based on the ambient (output) space, and the prior distribution $p(z)$ is based on the latent space.

From the perspective of CFP theory, the two entities *A* and *B* in CFP correspond to the high-dimensional data structure in the ambient (input) space and the low-dimensional data structure in the latent space, respectively. Based on this interpretation, the newly generated data structure in the ambient (output) space—produced by the generator—can be regarded as crSTR, formed by the convergence and fusion of *A* and *B* in CFP theory. Moreover, as outlined in section 2, the use of the inverse function of the Jacobian to express the qualitative expansion of relative space-time can be interpreted as crSTR induced by the mechanistic structure of DCPSs in CFP theory. This is because, from the CFP perspective, both the use of the inverse function of the Jacobian and the posterior distribution involves geometrically creating crSTR. Consequently, this understanding clarifies why employing the pull-back metric as a way of bringing the Riemannian metric set in the ambient (input) space to the latent space produces an effect similar to utilizing the posterior distribution.

## 2) Observation of Density Distribution

### A. Learning Riemannian Metrics in Ambient Space

Since (Arvanitidis et al. 2020) establishes the Riemannian metric in the ambient (input) space, it enables the measurement of data density and curvature by learning the Riemannian metric in the ambient space. This allows the effective utilization of high-dimensional data information present in the ambient (input) space.

The method for measuring data density and curvature by learning the Riemannian metric is defined as follows:

$$M_\chi(x) = (\alpha \cdot h(x) + \epsilon)^{-1} \cdot \mathbb{I}_D \qquad (7)$$

where $h(x): \mathbb{R}^D \to \mathbb{R} > 0$, and $x$ is near the data manifold, h(x)→1, otherwise h(x)→0.

At this time, one of the effective and simple approaches in relation to $h(x)$ is to define $h(x) = w^T \phi(x)$ using a positive Radial Basis Function (RBF) network (Que and Belkin 2016). Additionally, $w \in \mathbb{R}^k_{>0}$ and $\phi_k(x) = \exp(-0.5 \cdot \lambda_k \cdot \|x - x_k\|_2^2)$ with bandwidth $\lambda_k > 0$. Also, $\alpha$ and $\epsilon > 0$ is scaling factors that setting the lower and upper limits of the metric, respectively.

In summary, using RBF network is a key method for measuring information on data density and curvature. The bandwidth $\sigma$ of the RBF network ($\lambda_k$ in Eq. (6) above) is used as the $\sigma^2$(variance) in the pull-back metric through the inverse function of the Jacobian.

### B. CFP Theory-Based Interpretation

On the other hand, if we understand the process of measuring the information on data density and curvature from the aspect of the Riemannian metric as a learning process, as shown above, the viewpoint of CFP theory can provide us with insight into the geometric structure of the data. This is because, by means of an RBF network, can be interpreted as a train of microscopic continuous developmental processes of crSTR from the lens of CFP theory (Fig. 7). Through this process, the data are related to each other and proceed to be structured into a higher dimension.

$$h(x) = w^T \exp(-0.5 \cdot \lambda_k \cdot \|x - x_k\|_2^2) \tag{8}$$

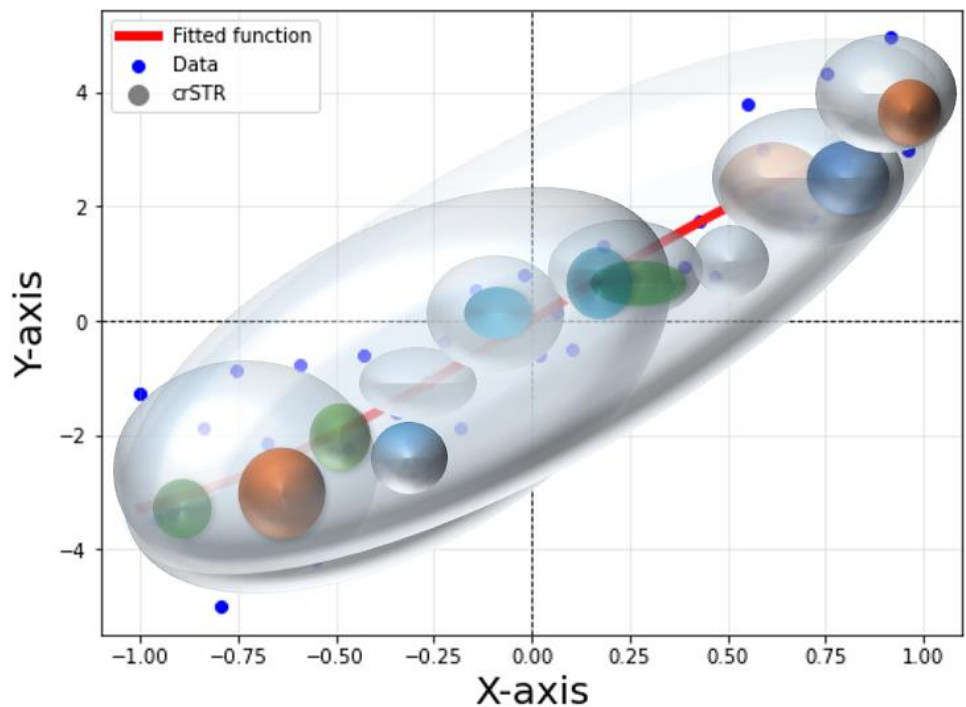

Figure 7. Understanding RBF Network in the Coordinate Plane regarding CFP theory.

In particular, the most important factors in measuring the information on data density and curvature are the distance $\|x - x_k\|_2^2$ between data points and the bandwidth $\lambda_k$. The combination of these two elements ultimately determines the shape and direction of data density and curvature. From the perspective of CFP theory, the combination of these two elements can be understood as crSTR, in which parties *A* and *B* create a CFP relationship. Since this crSTR always creates the shortest distance, the chain of crSTR-clustering is made possible through this shortest distance[16].

Meanwhile, such crSTR is captured in the overlapped appearance by means of a microscopic perspective in which the ambient (input) space and latent space within VAEs are grasped on the premise of a path unit, which is the same as crSTR by the CFP theory in Section III.3.1)B.

---

[16] On the other hand, the CFP theory provides a multi-dimensional perspective on the optimization process, rather than a linear perspective. Therefore, the shortest distance between data calculated through the optimization process in data science can be understood as the creation of crSTR. More details will be provided in a forthcoming paper.

**C. Analysis of the Generator's side through the CFP theory**

When looking at the Generator from the perspective of the CFP theory, it is understood as a process of integrating the ambient (input) space and latent space through the pull-back metric and processing it into the ambient (output) space[17]. However, there is a fact that should be noted here. If the process of pulling back higher-level information through the ambient (input) space metric using the pull-back metric has an effect similar to using the posterior rather than the prior, the clusters of data generated by the generator function in the ambient (output) space are bound to be different from those in the ambient (input) space connected to the latent space.

This discrepancy arises due to the fact that, in contrast to the conventional Auto Encoders, which involve data compression and subsequent restoration and reconstruction without modification, this process must converge and fuse the data clusters in the ambient (input) space and those in the latent space into a unified entity in the ambient (output) space to generate a new data cluster. In this regard, Auto Encoders (AEs) naturally evolves into a VAEs as they employ more intricate and higher-dimensional data clusters.

Based on the above, the implicit understanding of pullback metric method, which assumes that the inherent high-level information of the ambient (input) space is virtually the same as the ambient (output) space, is bound to have subtle errors.

Upon closer observation, these errors are effectively absorbed into the region of covariance $\sigma$ in the prior distribution $p(z)$, serving as an error-handling mechanism within the pull-back matrix. This study argues, on the basis of Arvanitidis et al. (2020), that the difference between the data cluster in the ambient (input) space and the newly generated data cluster in the ambient (output) space is evidence of a truly unique and mystique process in which genuine creation occurs.

**4. Analysis of (Arvanitidis et al. 2021)**

---

[17] The fundamental reason for the pull-back metric, which brings the ambient metric into the latent space, integrating the ambient (input) space and the latent space, is that the ambient (input) space, the latent space, and the ambient (output) space, all of three spaces are all overlapped.

(Arvanitidis et al. 2021) utilizes a prior to approximate the Riemannian metric in the latent space. The prior $p(z)$ represents an expectation of how latent variables are distributed in the latent space and is a predefined assumption before the model is trained. This prior determines the structure of the latent space and allows the generator to operate based on this structure when transforming latent variables into the ambient (output) space. Since (Arvanitidis et al. 2021) employs prior probability to approximate the Riemannian metric in the latent space, it can be seen as more similar to (Arvanitidis et al. 2018) than to (Arvanitidis et al. 2020), where the Riemannian metric is set up and used in the ambient (input) space.

An essential distinction between (Arvanitidis et al. 2018) and (Arvanitidis et al. 2021) is that in the latter, the prior probability distribution plays a critical role in constructing the Riemannian metric. This distinction defines the boundary between the two studies. In other words, (Arvanitidis et al. 2018) assumes that the variance $\sigma^2$ is uniformly distributed in the prior probability distribution, making it necessary to precisely measure the information related to data density beyond this assumption. To achieve this, they employ RBF networks in the latent space and integrate them into the pull-back Riemannian metric based on the prior probability distribution. In contrast, (Arvanitidis et al. 2021) directly incorporates the information related to data density into the latent space through the prior probability distribution.

The key question is how to reflect higher-level information, such as data density, directly through the prior distribution in the latent space without using an RBF network. To address this, (Arvanitidis et al. 2021) introduces two novel methodologies:

1. Employing an Energy-Based Model (EBM): This establishes a new prior-based Riemannian metric in the latent space.
2. Utilizing the Locally Conformally Flat Property (Conformal Property): This enables direct measurement of the metric within the latent space.

These two methodologies are intricately interconnected, forming the foundation for integrating data density into the prior probability distribution within the latent space.

**1) Locally Conformally Flat Riemannian Metric**

Firstly, this study reviews the process of setting a new prior-based Riemannian metric in the latent space using EBM. The new locally conformally flat Riemannian metric is given by:

$$M_\psi(z) = m(z) \cdot \mathbb{I}_d = \left(\alpha \cdot \nu_\psi(z) + \beta\right)^{-2/d} \cdot \mathbb{I}_d \tag{9}$$

where $\mathbb{I}_d$ the d-dimensional identity matrix, $\alpha$ and $\beta$ the positive scaling constant which controls lower and upper bound.

The function $m(z)$ scales the metric according to the location, transforming the original metric into a form where only its magnitude varies depending on the spatial position. This transformation ensures the conservation of energy during spatial transformations, leading to the metric being considered conformally flat. Specifically, Euclidean metric $\mathbb{I}_d$ is used in the latent space by default, where distances between all points are computed in an identical manner. Using the conformal property, distance measurement can be adjusted. In high-density data regions where $v_{\psi(z)}$ is large, $m(z)$ decreases, reducing $M_{\psi(z)}$ and shortening distances between data points, making them appear more closely connected. Conversely, in low-density regions where $v_{\psi(z)}$ is small, $m(z)$ increases, enlarging $M_{\psi(z)}$ and increasing distances, making data appear farther apart. This conformally flat metric thus enables more precise distance adjustments in the latent space, enhancing structural data representation.

Next, an understanding of $v_{\psi(z)}$, which serves as a learnable prior providing data density information in the latent space, is necessary. It is defined as:

$$v_\psi(z) = \frac{\exp(f_\psi(z))p(z)}{C} \tag{10}$$

Expanding Eq. (10), the interpretation is as follows: First, the energy function $f_{\psi(z)}$ is computed for latent variable $z$, where energy is inversely proportional to data density, leading to lower energy in densely populated areas. Then, the exponential function of $f_{\psi(z)}$ is combined with the prior $p(z)$ to define a new distribution, $v_\psi(z)$, thereby adjusting the distribution such that denser regions have greater values. Finally, the normalization constant $C$ ensures that the distribution integrates to 1.

As evident in Eq. (10), $f_\psi(z)$ assigns energy to data in the latent space, allowing structural representation. Due to energy properties, even when assigned in the latent space, a global understanding of data structure remains feasible, accommodating various characteristics and uncertainties. Thus, unlike (Arvanitidis et al. 2020), which required separate density measurements in the ambient space for higher-level information, (Arvanitidis et al. 2021) eliminates this necessity through EBM, its defining feature.

**2) A Learnable Prior for VAEs**

## A. Introduction to the Noise Contrastive Prior (NCP)

To further explore the methodology employed in (Arvanitidis et al. 2021), we must examine how its learnable prior adapts through training. The paper utilizes the "Noise Contrastive Prior (NCP)," as introduced by (Aneja et al. 2020).

According to (Aneja et al. 2020), NCP-VAEs training consists of two main stages. First, standard VAE training optimizes the base prior $p(z)$, aligning it as closely as possible with the posterior $p(z|x)$ (Stage 1 – VAE training). Next, the re-weighting function $r(z)$ is learned using Noise Contrastive Estimation (NCE), enhancing prior flexibility and posterior alignment (Stage 2 – NCE training).

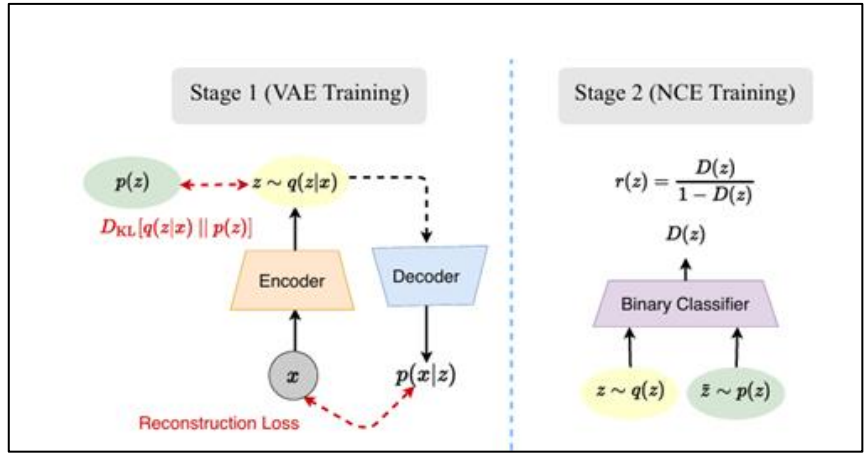

Figure 9. NCP-VAEs trained in two stages. The image is adapted from (Aneja et al. 2020).

NCE trains a binary classifier $D(z)$ by sampling from both posterior $q(z)$ and prior $p(z)$, distinguishing between the two. This is achieved by minimizing the binary cross-entropy loss:

$$\min_{D} -\mathbb{E}_{z \sim q(z)}[\log D(z)] - \mathbb{E}_{z \sim p(z)}[\log(1 - D(z))], \qquad (11)$$

where a binary classifier $D(z)$ that distinguishes samples from the posterior and the prior, and through this classifier, the re-weighting function $r(z)$ can be approximated in the following form under the optimized condition:

$$r(z) \approx \frac{D^*(z)}{1 - D^*(z)} \qquad (12)$$

This re-weighting function (Eq. (12)) is trained to estimate the ratio between the posterior and prior well, making the prior of the VAEs more flexible and effective.

Through this approach, NCP-VAEs compare noise samples with actual data to determine whether specific latent space regions contain meaningful information. Unlike conventional extrapolation techniques aimed at minimizing posterior-prior discrepancies, this method actively reduces prior probability in meaningless regions of the latent space. As a result, NCP-VAEs mitigate extrapolation issues in latent code-absent regions, meaningfully controlling posterior-prior discrepancies. Moreover, the optimization process minimizing NCP energy implicitly reduces probability density in unobserved regions, similar to normalization constants. This implicit normalization technique adjusts latent space probabilities without explicitly computing normalization constants, significantly reducing high-level space normalization complexity and computational costs.

In summary, the process of bringing high-level information inherent in the ambient (input) space, through pull-back metric, into the latent space in (Arvanitidis et al. 2020) can be simply replaced with the process of energy optimization in EBM.

## B. Approximation of Posterior $p(z|x)$ in NCP-VAEs, its Limitations and Contradictions

Originally, posterior $p(z|x)$ is a variable $z$-distribution in latent space for real data $x$. The encoder network of VAEs approximates a complex function that maps the input data $x$ to the latent space $Z$ using a DNNs. This function maps the input data to a specific location in the latent space, which is expressed as the mean and variance. However, it is important to note that the posterior $p(z|x)$ cannot be obtained exactly, so instead, an approximation based on the prior distribution is used, and in the process, the KL-divergence is minimized to make it similar to the prior distribution $p(z)$. As a result, in the process of approximating with DNNs, there happens a problem of arbitrarily extrapolating areas in the latent space where the data does not exist, based on the prior distribution. This process can cause the prediction of meaningless probability distributions, ultimately causing problems with the model's reliability and consistency.

This problem of arbitrary extrapolation can also affect NCP-VAEs. Even with NCP-VAEs, the data distribution is not identical to the posterior $p(z|x)$ because it is approximated via prior distribution $p(z)$ in the DNN. Thus, extrapolation may still not be entirely resolved, and the problem of not accurately obtaining the posterior distribution may persist. Additionally, given that the goal of NCP-VAEs is to make the prior $p(z)$ closely approximate the posterior $p(z|x)$, bringing the two prior distributions into closer alignment rather than accurately estimating posterior characteristics results in a fundamental contradiction. This contradiction raises concerns about the reliability of NCP-VAEs, as the process ultimately reduces to making prior and posterior distributions appear similar rather than providing a more accurate posterior estimate.

In other words, the data distribution sample $p(z)$ used in NCP-VAEs is fundamentally merely the prior distribution learned through a DNNs, so it cannot be considered a completely accurate posterior distribution. Of course, as we have seen earlier, NCP-VAEs learn to reduce the prior probability in the meaningless region of the latent space, so it can be seen as controlling the learning more meaningfully than simply performing extrapolation to minimize the difference between the posterior and prior (Aneja et al. 2020). Nevertheless, the data sample is still only an approximate posterior distribution calculated by DNNs based on the prior distribution, and there is still uncertainty about the areas without latent codes[18].

### 3) Summary and evaluation

(Arvanitidis et al. 2021) identified several fatal practical drawbacks of (Arvanitidis et al. 2020) – a. the difficulty in selecting the number of components $K$ as well as their parameters as the bandwidth $\sigma$ in modelling the precision with an RBF; b. the non-robustness of metrics due to the curse of dimensionality in the latent space; and c. the problem of the computational cost of the Jacobian and its derivatives – and they proposed a new locally conformally flat Riemannian metric as a simple, efficient, and robust alternative to pullback metrics. It learns based on prior distributions using EBM, which can grasp the structure of the entire data even if energy is allocated in the latent space, so it does not need to include high-level information by measuring data density using RBF separately in the ambient (input) space such as (Arvanitidis et al. 2020).

Despite these advancements, a fundamental limitation remains in the approach taken by (Arvanitidis et al. 2021). Although NCP-VAEs mitigate extrapolation by lowering prior probability in regions of the latent space that do not correspond to meaningful data, the prior distribution samples $p(z)$ still ultimately rely on approximations made by DNNs. This means that the learned distribution is not an exact representation of the posterior $p(z|x)$. While NCP-VAEs adjust for this issue more effectively than simple extrapolation techniques, the underlying problem persists.

Furthermore, given that the primary goal of NCP-VAEs is to approximate the prior $p(z)$ as closely as possible to the posterior $p(z|x)$, it raises a conceptual contradiction: if the prior $p(z)$ is merely an approximation of $p(z|x)$ using DNNs, then the process essentially consists of making two approximations as similar as possible rather than accurately modeling the posterior distribution itself. This contradiction weakens the theoretical foundation of NCP-VAEs as a truly robust alternative to previous methods.

---

[18] KL-divergence has been shown to reduce the discrepancy between the posterior and prior distributions (Bishop 2006). However, if the posterior distribution itself is not accurately modelled, minimizing KL-divergence may lead to meaningless results. This is because, unlike representing probabilities in the compressed, two-dimensional basis of the latent space, the posterior distribution requires representing probabilities in the multi-dimensional basis of the ambient input space in order to obtain high-level information. However, this structural difference is ignored in KL-divergence minimization, which assumes a latent space. CFP theory provides a perspective that makes these differences explicit.

In summary, the locally conformally flat Riemannian metric proposed in (Arvanitidis et al. 2021) presents a computationally efficient alternative to the pull-back metric in (Arvanitidis et al. 2018), eliminating the need for additional high-level information measurement in the ambient space. However, its reliance on approximated priors still introduces uncertainty in posterior estimation, making its overall improvement in performance less significant than initially expected. This underscores the need for further refinement in generative modeling techniques to address the limitations in latent space structuring and posterior distribution accuracy.

## IV. Future Research Tasks

### 1. Utilization of Hilbert Space

**1) Calculation of Posterior Distribution and Dilemma of Selection**

Meanwhile, what becomes clear from the analysis of (Arvanitidis et al. 2021) as well as (Arvanitidis et al. 2018) and (Arvanitidis et al. 2020) is that the limitation in calculating outcomes using VAEs stems from the fact that the posterior $p(z|x)$ cannot be mathematically obtained in practice. The reason why posterior $p(z|x)$ is crucial is that if the difference between the data clusters in the ambient (input) space and the new data clusters in the ambient (output) space—essentially a kind of gap—is the evidence of spontaneous and active data generation, then it can be said that this generation should begin with the posterior $p(z|x)$. Therefore, approximating or capturing the posterior $p(z|x)$ is a fundamental task in understanding data generation via VAEs and in establishing a mechanism to generate data accurately.

Regarding the methods to identify the posterior $p(z|x)$, (Arvanitidis et al. 2020) employs the inverse function of Jacobian as a substitute for the posterior $p(z|x)$, while (Arvanitidis et al. 2021) applies an EBM-based NCP-VAE to indirectly calculate posterior $p(z|x)$ through the prior distribution.

To elaborate further, (Arvanitidis et al. 2020) can incorporate high-level information from the ambient (input) space into the latent space by making use of its unique way. When the inverse function of the Jacobian is applied in the latent space after setting the Riemannian metric in the ambient (input) space, the ambient metric placed in the ambient (input) space is pulled into the latent space via the Riemannian metric $M_\chi\big(\mu(z) + \mathrm{diag}(\epsilon) \cdot \sigma(z)\big)$. This allows the inclusion of high-level information from the ambient (input) space. Consequently, even though the DNNs still utilize prior $p(z)$ as an approximation function in the latent space, calculating the pull-back metric can yield the same effect as using posterior $p(z|x)$ instead of prior $p(z)$. Nevertheless, this approach suffers from the aforementioned critical practical drawbacks, particularly excessive computational constraints and costs.

In contrast, (Arvanitidis et al. 2021) employs NCP-VAEs, which does not explicitly require posterior $p(z|x)$ by learning to reduce the prior probability in meaningless regions of the latent space instead of extrapolating. However, as previously discussed, the data distribution sample $p(z)$ utilized in NCP-VAEs is fundamentally an approximation of prior $p(z)$ via DNNs, making it an indirect replacement for posterior $p(z|x)$. This issue presents a circular logical flaw and leaves open the possibility that the extrapolation problem has not been entirely resolved. Consequently, this limitation significantly impacts the validity of VAEs-based outcome generation.

In summary, the approach in (Arvanitidis et al. 2020) of defining a pull-back metric through the inverse function of the Jacobian effectively implements the Riemannian metric as the complete crSTR of CFP theory by inserting reversed time via the dimensional transcendence property inherent in mathematical matrices and vector inner products, and the overlapping phenomenon present in VAEs. Conversely, (Arvanitidis et al. 2021) attempts to bypass the computational costs associated with the Jacobian's inverse function in (Arvanitidis et al. 2020) by obtaining the posterior distribution more efficiently in the latent space via EBM. However, the process of replacing probability distributions through a DNNs universal probability approximator introduces a significant logical flaw. Ultimately, the contrast between (Arvanitidis et al. 2020) and (Arvanitidis et al. 2021) can be condensed into a debate of "methodological logical completeness vs. technical feasibility."

**2) Appearance of Hilbert Space as Alternative**

Meanwhile, even though the process of setting a pull-back metric through the Jacobian's inverse function (Arvanitidis et al. 2020) can be understood within Euclidean space, these processes actually unfold under conditions that go beyond Euclidean space. Given this, it is essential to explore a spatial framework that overcomes the constraints of Euclidean space. It is evident that the physical location where these conditions materialize should exist within the internal spatial structure of VAEs, allowing for overlap. Based on this point, this study aims to reinterpret the internal spatial structure of VAEs by transitioning from Euclidean space to Hilbert space.

**A. Differences from Euclidean Space – Especially Regarding Jacobian**

Hilbert space is a complete vector space with a defined inner product, extending beyond finite dimensions to infinite dimensions (Conway 2019; Rudin 2012). This space reflects high-dimensional properties of data and offers the potential to handle non-Euclidean characteristics. These properties also influence how the Jacobian operator functions in Hilbert space (Conway 2019). Fundamentally, the Jacobian in Hilbert space expresses the gradient of a linear transformation, as it does in Euclidean space, but in infinite-dimensional Hilbert space, it is defined as a linear operator rather than a matrix representation (Lang 2006). Thus, from a computational perspective, the Jacobian in Hilbert space can better reflect complex high-dimensional data interactions while maintaining linearity, owing to its inner product-based structure. This makes it particularly useful for capturing nonlinear structures in data and defining pull-back Riemannian metrics in the latent space.

**B. Hilbert Space and Possibility of Capturing Generation Process**

Furthermore, a crucial question arises: Can the inner product and completeness property of Hilbert space capture the process of generating new data clusters rather than merely compressing and restoring data in the generator? In other words, how effectively can Hilbert space represent the convergence and fusion of data in the ambient (input) space and the latent space into a new data cluster in the ambient (output) space compared to Euclidean space?

From the CFP theory perspective, while the structure of "Duplexity-Contradiction" can be represented in Euclidean space, the dynamic process of transforming from "Duplexity-Contradiction" to "Paradox-Stratification"—creating new manifolds through thorough closure and eternal openness—cannot be fully expressed. Furthermore, Euclidean space imposes a constraint of differentiability, reducing "Duplexity-Contradiction" to "reflexive manifestation of ontological unit," thus limiting dimensional transformation to a linear expression of "quantitative expansion" rather than "generation accompanied by qualitative transformation."

In contrast, Hilbert space, by leveraging "high-dimensional inner products" and "matrix operations," allows for richer interaction modeling between data, enabling the generation of new data rather than mere restoration. More specifically, Euclidean space typically represents matrices in a diagonalized form based on the coordinate system at a given time. This diagonalization effectively represents dimensions, but the interactions captured in off-diagonal components are often underrepresented. Unlike Euclidean space, which constrains interactions to diagonalized forms, Hilbert space provides a more flexible framework where both diagonal and non-diagonal terms contribute to the representation of data relationships.

By applying this perspective, the process of merging data clusters in the latent space with those in the ambient (output) space can be expressed more naturally in Hilbert space. The non-diagonal terms and completeness theorem of generalized Hilbert space facilitate a more detailed representation of data fusion and new cluster generation. When representing high-dimensional generated data, the Cauchy sequence converges at both extremes of Hilbert space, ensuring consistency in data transformations and enabling a stable linear transformation process.

**C. Limitations – Quantitative Approximation of Qualitative Processes due to Linearity**

How should Hilbert space be understood from the perspective of CFP theory? Unlike Euclidean space, can Hilbert space dynamically express the structure of 'Paradox - Stratification' beyond the structure of 'Duplexity - Contradiction'?

The ability of Hilbert space to express data interactions embedded in non-diagonal components of a matrix suggests that it is a step closer to the transformation into a 'Paradox - Stratification' structure. In this sense, Hilbert space allows for a better representation of the relative space-time creation process, which is inserted through the reversed time in VAEs, a concept that Euclidean space fails to fully capture. However, Hilbert space also carries inherent constraints, such as the convergence condition of the Cauchy sequence and the linearity of matrix operations.

The completeness condition, which requires all Cauchy sequences to converge, ensures that new data generation remains stable and consistent. However, this stability comes at the cost of restricting the degree of freedom in the data generation process, as data should remain within predefined mathematical boundaries. Additionally, while the linearity of matrix operations allows Hilbert space to represent interactions that Euclidean space struggles to express, it also imposes a fundamental constraint—preventing the complete capture of complex nonlinear interactions between data. This suggests that the qualitative generation process of data can only occur within a larger structural framework.

In other words, the qualitative generation process—defined as the convergence and fusion of data—is still merely being approximated through a more refined and expansive quantitative measurement unit. To illustrate this concept metaphorically, this is akin to creating the animation from a sequence of images: By using a larger space and more images, the animation appears smoother and more dynamic, but the images themselves do not become a real animation. Similarly, Hilbert space allows for a richer representation of paradoxical dimensional structures than Euclidean space, yet it still falls short of achieving the complete realization of nonlinearity and qualitative data creation processes.

In this context, quantum state space, as studied in quantum mechanics, is a physical adaptation of Hilbert space axioms that enables the handling of nonlinear interactions through quantum operators functioning as continuous transformations. In this framework, superposition and entanglement emerge as essential concepts in processing information, and since generated data clusters also constitute a form of information, they might exhibit similar properties. This offers a physically analogous approach to modeling data superposition in VAEs, aligning with the structure outlined in previous sections. Consequently, the Hilbert space in quantum mechanics has the potential to accommodate more intricate nonlinear interactions than standard Hilbert spaces.

## 2. Theory of Passivity as Intersection Point between Linear Approximation and Posterior Distribution

A fundamental question remains: Is it possible to develop a mathematical framework that nonlinearly expresses the qualitative generation process of data, rather than relying on a quantitative approximation through linear methods?

Up to this point, qualitative aspects of data generation that necessitate nonlinear representation have typically been approximated using mathematical tools, particularly through probabilistic methods such as Bayesian inference. The core mathematical methodology of (Arvanitidis et al. 2020)—which establishes a connection between linear approximation and posterior $p(x|z)$ via pull-back metric—reinforces this claim. However, such instrumental methodologies are inherently a posteriori rather than fundamental, and no existing mathematical framework directly expresses qualitative transitions in their actual state of occurrence.[19]

To overcome these limitations, this study proposes the theory of passivity in CFP theory, also known as "Leave As It Is (LAI)," as a philosophical supplement. In CFP theory, passivity represents the silent condition necessary for establishing 'Thorough Closure – Eternal Opening (TC-EO),' which is the process of reaching infinite openness through absolute closure. As discussed earlier (See Section II. 2. 2.), this transformation should accompany the transition from 'Duplexity - Contradiction' to 'Paradox - Stratification'. This process does not actively reveal itself but instead unfolds naturally over time, manifesting passively as an emergent phenomenon.

The theory of passivity posits that when thorough convergence surpasses a critical point, a fusion occurs that opens a novel unexpected dimension. The assumption of fusion at this moment signifies the operation of passivity. While this concept extends beyond mathematical rigor, it appears viable in practical applications such as data science, where real-world contexts necessitate empirical adaptations. This is because real data originates from the physical world rather than an abstract mathematical construct.

In essence, passivity functions as a bridge between abstract mathematical limitations and real-world data generation by integrating philosophical reasoning into computational frameworks. This conceptual augmentation is particularly relevant for data science applications, where real data is influenced by complex interactions and contextual dependencies that cannot be entirely encapsulated by mathematical models. Consequently, the theory of passivity provides an essential philosophical foundation to support the unpredictable emergence of novel data structures beyond what current mathematical models can

---

[19] To overcome this limitation, (Arvanitidis et al. 2021) appears to employ a physical methodology in addition to mathematical tools. This is supported by the fact that they used an Energy-Based Model (EBM).

explicitly define.[20]

## VI. Conclusion

So far, we have examined how the data generation process within VAEs, a type of generative model, can be captured through the lens of CFP theory. The conventional understanding of VAEs is that existing data is transformed and restored in the ambient (output) space by compressing the ambient (input) space into the latent space and replacing the posterior $p(z|x)$ with a variational $q(z|x)$ in the latent space. In contrast, when VAEs are understood from the perspective of CFP theory, both the ambient (input) space and the compressed latent space can be assumed to be the existence $A$ and the recursive existence $B$ of that existence $A$. Further, the crSTR, in which the existence $A$ and its recursive existence $B$ are converged and fused, can be considered as the ambient (output) space.

This CFP-theoretical perspective provides an opportunity to gain insight into the overlap between the ambient (input) space, the ambient (output) space, and the latent space. Understanding this overlap is significant because it provides a foundation for understanding the organic relationship between low-dimensional and high-dimensional data. The overlap between the ambient (input) space and the latent space can be understood as a process in which an existence $A$ transforms into a compressed form of itself as existence $B$. Since the existence $B$, which is recursively derived from the existence $A$, is a generative body that has undergone compression, it can be interpreted as a quantitative dimensional reduction. This perspective enables the establishment of a Riemannian metric based on Euclidean space (Arvanitidis et al. 2020) (Ambient metric).

In contrast, the overlap between the latent space and the ambient (output) space can be understood as the superposition between the recursive existence $B$ and the crSTR which serves as the ambient (output) space. Therefore, the specific appearance of this superposition cannot be understood merely as a quantitative expansion. This is because crSTR, as a relative space-time, arises through the relationships between the entities involved, thereby necessitating qualitative transformation. For this reason, the specific formation of the ambient (output) space should be viewed as a process of creation, rather than simply as modified restoration or reconstruction. Since this creation space cannot be fully described using the conventional Euclidean framework, a new spatial metric is required.

The pull-back metric established based on the inverse function of the Jacobian in the latent space in (Arvanitidis et al. 2020) actually supports the necessity of this new spatial metric. The Jacobian inverse function, which presupposes overlap as a fundamental property of VAEs, inserts reversed time, transforming the Euclidean understanding of spatial expansion—which is limited to quantitative growth—into qualitative expansion within relative space-time. When examined through the lens of CFP theory, such qualitative expansion constitutes a generative domain, which emerges through the formation process of crSTR. However, this generative domain remains unmeasurable within the conventional Euclidean metric framework.

---

[20] A detailed explanation of this is beyond the scope of this study and will be addressed in greater detail in a subsequent analysis.

  This study explores the potential of Hilbert space as an alternative spatial metric capable of accommodating qualitative expansion within relative space-time. Since Hilbert space reflects the high-dimensional nature of data and provides a framework for handling non-Euclidean properties, it allows the convergence and fusion of data from the ambient (input) space and the latent space into new data clusters in the ambient (output) space. This process can be expressed more naturally and comprehensively in Hilbert space than in Euclidean space. However, due to the convergence condition of the Cauchy sequence and the inherent linearity of matrix operations, Hilbert space still approximates the qualitative generative process of data fusion and convergence through denser and expanded quantitative measurements rather than fully realizing a purely qualitative transformation. Therefore, this study proposes that overcoming these limitations requires philosophical supplementation, and it introduces CFP theory's concept of passivity as a hypothesis to address this challenge.